\documentclass{article}

\usepackage[preprint]{neurips_2025}

\usepackage[utf8]{inputenc} % allow utf-8 input
\usepackage[T1]{fontenc}    % use 8-bit T1 fonts
\usepackage{hyperref}       % hyperlinks
\usepackage{url}            % simple URL typesetting
\usepackage{booktabs}       % professional-quality tables
\usepackage{amsfonts}       % blackboard math symbols
\usepackage{nicefrac}       % compact symbols for 1/2, etc.
\usepackage{microtype}      % microtypography
\usepackage{xspace}
\usepackage{microtype}
\usepackage{graphicx}
\usepackage{caption} %% used by snippet-styling
\usepackage{mdframed} %% used by snippet-styling
\usepackage{subfigure}
\usepackage{multirow}
\usepackage[table,xcdraw]{xcolor}
\usepackage{amsmath}
\usepackage{cleveref}
\usepackage{listings}
\usepackage{xcolor} % colors defined in macros.tex

\usepackage{pifont}
\usepackage{enumitem}
\title{MetaGen: A DSL, Database, and Benchmark for VLM-Assisted Metamaterial Generation}

\author{%
  Liane Makatura\thanks{Equal contribution} \\
  MIT CSAIL
  \And
  Benjamin Jones$^{*}$ \\
  MIT CSAIL
  \And 
  Siyuan Bian \\
  Shanghai Jiao Tong University
  \And
  Wojciech Matusik \\
  MIT CSAIL
}

\begin{document}

%% ==== Author comment commands ====

% Author Comment Macros
\newif\ifshowauthorcomments
\showauthorcommentstrue
% \showauthorcommentsfalse
\newcommand{\authorcomment}[3]{\ifshowauthorcomments{\bfseries \scriptsize \color{#3} #1: #2}\fi}
\newcommand{\ben}[1]{\authorcomment{BJ}{#1}{violet}}
\newcommand{\liane}[1]{\authorcomment{LM}{#1}{red}}
\newcommand{\wojciech}[1]{\authorcomment{WM}{#1}{blue}}

\newcommand{\ignore}[1]{}
\newcommand{\todo}[1]{\textcolor{red}{\textbf{#1}}}
\newcommand{\toverify}[1]{\textcolor{red}{\textbf{#1}}}
\newcommand{\tofill}[1]{\textcolor{red}{\textbf{#1}}}
\newcommand{\torevisit}[2]{\textcolor{purple}{ \textbf{REVISIT({#2}): }{#1}}}

\newenvironment{freeze}
    {
        \textcolor{red}{\textbf{IMPORTANT: Liane is editing the greyed-out text offline -- check in before adjusting to prevent duplicated energy / lost changes}}
        \par\color{lightgray}
    }
    {\par}

%% ==== Latin phrases and abbreviations ====
% Add a period to the end of an abbreviation unless there's one
% already, then \xspace. Requires \usepackage{xspace}
\makeatletter
\DeclareRobustCommand\onedot{\futurelet\@let@token\@onedot}
\def\@onedot{\ifx\@let@token.\else.\null\fi\xspace}

\def\eg{e.g\onedot} \def\Eg{E.g\onedot}
\def\ie{i.e\onedot} \def\Ie{I.e\onedot}
\def\cf{c.f\onedot} \def\Cf{C.f\onedot}
\def\etc{etc\onedot} \def\vs{vs\onedot}
\def\wrt{w.r.t\onedot} \def\dof{d.o.f\onedot} \def\WLOG{w.l.o.g\onedot}
\def\etal{et al\onedot}
\makeatother

%% ======== general math commands =========
\newcommand{\R}{\mathbb{R}}

%% ========= paper specific notation ==========
\newcommand{\youngsMod}{E}
\newcommand{\poissonRatio}{\nu}
\newcommand{\density}{\rho}

\newcommand{\tInterp}{$\vec{t}$\xspace}
\newcommand{\thickness}{$t$\xspace}
\newcommand{\procMeta}{ProcMeta\xspace}

%% ========= LLM name formatting ==========
\newcommand{\novalite}{NovaLite}
\newcommand{\llava}{LLaVA\xspace}
\newcommand{\ohOne}{O1\xspace}
\newcommand{\fourOhMini}{GPT-4o-mini\xspace}

%% ========= DSL element formatting ==========
\newcommand{\dslFrag}[1]{\texttt{#1}}

%% ========= values/terms that might change ==========
\newcommand{\numMaterialsInDatabase}{$153,263$\xspace}
\newcommand{\numMaterialsInDatabaseRoundedDown}{$150\,000$\xspace}

%% Requires \usepackage{xcolor}
%% Requires \usepackage{lstlistings}
\definecolor{codegreen}{rgb}{0,0.6,0}
\definecolor{codeblue}{rgb}{.11,.56,1}
\definecolor{codegray}{rgb}{0.5,0.5,0.5}
\definecolor{codepurple}{rgb}{0.58,0,0.82}

\definecolor{codeKeyword}{RGB}{211	54	130}
% \definecolor{codeKeyword}{RGB}{133, 153, 0}
\definecolor{codeComment}{RGB}{42	161	152}
\definecolor{codeOmitted}{RGB}{108	113	196}
\definecolor{codeNumbers}{rgb}{0.5,0.5,0.5}
\definecolor{codeString}{RGB}{128, 161, 16}

\definecolor{textusercolor}{RGB}{40 20 10}
\definecolor{textgptcolor}{RGB}{62, 65, 115}

\definecolor{codebackcolour}{RGB}{	253	246	227}
\definecolor{backuserprompt}{RGB}{
253, 250, 250}
%\definecolor{backgptresponse}{rgb}{0.93,0.93,0.98}

\definecolor{backgptresponse}{RGB}{226 228 255}

%% Defining custom languages -- try not to put styling here, only keywords/comment signifiers/etc. this way all languages will look the same. 
\lstdefinelanguage{JavaScript}{
  keywords={typeof, new, true, false, catch, function, return, null, catch, switch, var, if, in, while, do, else, case, break, const},
  ndkeywords={class, export, boolean, throw, implements, import, this, require},
  sensitive=false,
  comment=[l]{//},
  morecomment=[s]{/*}{*/},
  morestring=[b]',
  morestring=[b]"
}

\colorlet{punct}{red!60!black}
\definecolor{delim}{RGB}{20,105,176}
\colorlet{numb}{magenta!60!black}
\lstdefinelanguage{json}{
    numbers=left,
    numberstyle=\scriptsize,
    stepnumber=1,
    numbersep=8pt,
    showstringspaces=false,
    breaklines=true,
    frame=lines,
    literate=
     *{0}{{{\color{numb}0}}}{1}
      {1}{{{\color{numb}1}}}{1}
      {2}{{{\color{numb}2}}}{1}
      {3}{{{\color{numb}3}}}{1}
      {4}{{{\color{numb}4}}}{1}
      {5}{{{\color{numb}5}}}{1}
      {6}{{{\color{numb}6}}}{1}
      {7}{{{\color{numb}7}}}{1}
      {8}{{{\color{numb}8}}}{1}
      {9}{{{\color{numb}9}}}{1}
      {:}{{{\color{punct}{:}}}}{1}
      {,}{{{\color{punct}{,}}}}{1}
      {\{}{{{\color{delim}{\{}}}}{1}
      {\}}{{{\color{delim}{\}}}}}{1}
      {[}{{{\color{delim}{[}}}}{1}
      {]}{{{\color{delim}{]}}}}{1},
}

% \lstset{
%   basicstyle=\ttfamily,
%   columns=fixed,
%   fontadjust=true,
%   basewidth=0.5em
% }

%% consistent styling used for all languages
\lstdefinestyle{codestyle}{
    backgroundcolor=\color{codebackcolour},
    commentstyle=\color{codeComment},
    keywordstyle=\color{codeKeyword},
    numberstyle=\tiny\color{codeNumbers},
    stringstyle=\color{codeString},
    basicstyle=\linespread{1}\footnotesize,
    columns=flexible,
    breakatwhitespace=false,         
    breaklines=true,                 
    captionpos=b,                    
    % numbers=left,                    
    % numbersep=5pt,     
    showspaces=false,
    showstringspaces=false,
    showtabs=false,
    tabsize=2,
    escapeinside={\$}{\$},
}
\lstset{style=codestyle}
%% Requires \usepackage{xcolor}
%% Requires \usepackage{lstlistings}
\definecolor{codegreen}{rgb}{0,0.6,0}
\definecolor{codeblue}{rgb}{.11,.56,1}
\definecolor{codegray}{rgb}{0.5,0.5,0.5}
\definecolor{codepurple}{rgb}{0.58,0,0.82}

\definecolor{codeKeyword}{RGB}{211	54	130}
% \definecolor{codeKeyword}{RGB}{133, 153, 0}
\definecolor{codeComment}{RGB}{42	161	152}
\definecolor{codeOmitted}{RGB}{108	113	196}
\definecolor{codeNumbers}{rgb}{0.5,0.5,0.5}
\definecolor{codeString}{RGB}{128, 161, 16}

\definecolor{textusercolor}{RGB}{40 20 10}
\definecolor{textgptcolor}{RGB}{62, 65, 115}

\definecolor{codebackcolour}{RGB}{	253	246	227}
\definecolor{backuserprompt}{RGB}{
253, 250, 250}
%\definecolor{backgptresponse}{rgb}{0.93,0.93,0.98}

\definecolor{backgptresponse}{RGB}{226 228 255}

\newcommand{\gpticon}{figures/formatting/chatgpt-logo.png}
\newcommand{\usericon}{figures/formatting/person-raising-hand.png}

%% Defining custom languages -- try not to put styling here, only keywords/comment signifiers/etc. this way all languages will look the same. 
\lstdefinelanguage{JavaScript}{
  keywords={typeof, new, true, false, catch, function, return, null, catch, switch, var, if, in, while, do, else, case, break, const},
  ndkeywords={class, export, boolean, throw, implements, import, this, require},
  sensitive=false,
  comment=[l]{//},
  morecomment=[s]{/*}{*/},
  morestring=[b]',
  morestring=[b]"
}

%% consistent styling used for all languages
\lstdefinestyle{codestyle}{
    commentstyle=\color{codeComment},
    keywordstyle=\color{codeKeyword},
    numberstyle=\tiny\color{codeNumbers},
    stringstyle=\color{codeString},
    basicstyle=\linespread{0.85}\footnotesize,
    columns=flexible,
    breakatwhitespace=false,         
    breaklines=true,                 
    captionpos=b,                    
    % numbers=left,                    
    % numbersep=5pt,     
    showspaces=false,
    showstringspaces=false,
    showtabs=false,
    tabsize=2,
    escapeinside={\$}{\$},
}

\surroundwithmdframed[
  hidealllines=true,
  backgroundcolor=codebackcolour,
  innerleftmargin=0pt,
  innertopmargin=0pt,
  innerbottommargin=0pt]{gptcodeblock}

\newcommand\colboxcolor{codeComment} % temporary
%% colored box -- takes background color as parameter
\newsavebox{\savedcolorbox}
\newenvironment{colbox}[2]
  {\renewcommand\colboxcolor{#1}%
   \begin{lrbox}{\savedcolorbox}%
    \begin{minipage}{\dimexpr\columnwidth-2\fboxsep\relax}

   \footnotesize
   \bgroup\color{#2}
   }
  {\egroup\end{minipage}\end{lrbox}%
   \begin{center}
   \colorbox{\colboxcolor}{\usebox{\savedcolorbox}}
   \end{center}
}

%% text responses
\newsavebox{\savedfigurebox}
\newenvironment{blurbwithfig}[5]
{
    \newcommand{\figurewidth}{#1}
    \newcommand{\iconwidth}{0.025\textwidth}
    \newcommand{\blurbwidth}{0.982\textwidth - \figurewidth - \iconwidth}
    \newcommand{\imagetoshow}{#2}
    \newcommand{\backgroundcolor}{#3}
    \newcommand{\boxtextcolor}{#4}
    \newcommand{\icontoshow}{#5}

    % collect the to-be-right-aligned figure block
    \begin{lrbox}{\savedfigurebox}%
    \begin{minipage}[t]{\figurewidth}
        \vspace{3pt}
        \ifthenelse{\equal{\imagetoshow}{}}{}{\includegraphics[width=\textwidth]{\imagetoshow}}
        % \captionof{figure}{note}
        % \label{fig:figure2}
    \end{minipage}\end{lrbox}%

    % start the user icon
    \noindent
    \begin{minipage}[t]{\iconwidth}
    \vspace{2pt}
    \centering
    \includegraphics[width=\textwidth]{\icontoshow}
    \end{minipage}
    %
    % start the left-aligned chat block
    \noindent
    \begin{minipage}[t]{\blurbwidth}
    \vspace{0pt}
    \begin{colbox}{\backgroundcolor}{\boxtextcolor}
}
{
    \end{colbox}
    \end{minipage}
    % end the primary prompt area
    \hfill
    % begin the image portion
    \usebox{\savedfigurebox}
}

\newenvironment{userprompt}[2]
{
    \begin{blurbwithfig}{#1}{#2}{backuserprompt}{textusercolor}{\usericon}
}
{
    \end{blurbwithfig}
}

\newenvironment{gptresponse}[2]
{
    \begin{blurbwithfig}{#1}{#2}{backgptresponse}{textgptcolor}{\gpticon}
}
{

    \end{blurbwithfig}
}

\lstnewenvironment{gptcodeblock}[1]
{
    \lstset{style=codestyle} %% change this to be a local setting, so it doesn't affect others
    \lstset{language=#1}
}
{}

\newenvironment{chat}[1]
{
    \newcommand{\preventbreaks}{#1}
    \begin{center}
    \mdfsetup{nobreak=\preventbreaks}
    \begin{mdframed}[
        linecolor=black,
        innerleftmargin=0.04cm,
        innerrightmargin=0cm
        innertopmargin=0cm
        innerbottommargin=0cm
    ]{}

}
{ 
    \end{mdframed}
    \end{center}
}

%% ============================
%% Macros to be used inside the chat environment, to help shorten/clarify the included chats
%% ============================

\newcommand{\authorremark}[1]{\footnotesize\textit{(Author remark: #1)}}

% parameters: #1 is the summary of the omitted content (will show up in doc if provided; can be left blank), #2 can be placed around the content to omit, so you can leave it in the tex, it just doesn't get rendered anywhere
\newcommand{\omitted}[2]{
    % \textcolor{codeblue}{
    \ifthenelse{\equal{#1}{}}{\textit{(... content omitted by authors ...)}} 
                            {\textit{(... omitted by authors: #1 ...)}}
}

% parameters: #1 is the summary of the omitted code (will show up in doc if provided; can be left blank), #2 can be placed around the content to omit, so you can leave it in the tex, it just doesn't get rendered anywhere
%% IMPORTANT: to use this inside a \gptcodeblock, you must wrap it in $$ to escape into latex mode. For example: 

% for x in range(0,3):
% $\omittedCode{update and output i}{
%.     i = x
%      print("the value of i is " + str(i))
% }$

\newcommand{\omittedCode}[2]{
    \textcolor{codeOmitted}{
    \ifthenelse{\equal{#1}{}}{\textit{(... code omitted by authors ...)}} 
                            {\textit{(... omitted by authors: #1 ...)}}
    }
}

\maketitle

\begin{abstract}

Metamaterials are micro-architected structures whose geometry imparts highly tunable—often counter-intuitive—bulk properties. Yet their design is difficult because of geometric complexity and a non-trivial mapping from architecture to behaviour. We address these challenges with three complementary contributions. \textbf{(i) \emph{MetaDSL}}: a compact, semantically rich domain-specific language that captures diverse metamaterial designs in a form that is both human-readable and machine-parsable. \textbf{(ii) \emph{MetaDB}}: a curated repository of more than \numMaterialsInDatabaseRoundedDown{} parameterized MetaDSL programs together with their derivatives—three-dimensional geometry, multi-view renderings, and simulated elastic properties. \textbf{(iii) \emph{MetaBench}}: benchmark suites that test three core capabilities of vision–language metamaterial assistants—structure reconstruction, property-driven inverse design, and performance prediction. We establish baselines by fine-tuning state-of-the-art vision–language models and deploy an omni-model within an interactive, CAD-like interface. Case studies show that our framework provides a strong first step toward integrated design and understanding of structure–representation–property relationships.

\end{abstract}

% ==================================================================================
% For full instructions, see the neurips_reference_material/neurips_2025.{tex, pdf} file
% ==================================================================================
% Summary:
%   - max 9 pages including figures. NOT COUNTED: references, checklist, appendices
%   - headings + figure/table captions should be all lower case (except first word and proper nouns)
%   - citations can be author/year or numeric (consistency is only requirement).  It is permissible to reduce the font size to small (9 point) when listing the references
%   - table: captions above table; no vertical rules
%   - PDF: US letter, watch out for fonts (reqs in neurips_2025.pdf)
%   - Checklist: see checklist.tex or neurips_2025.pdf

% ===========================
%      MAIN CONTENT
% ===========================

\section{Introduction}
\label{sec:intro}

Metamaterials have attracted intense research interest because microscale geometries can endow bulk matter with properties that are unattainable in the parent substance. Careful geometric tuning yields extraordinary behaviours such as programmable deformation\,\citep{Jennet_DiscreteMetamaterials_2020,babaee20133d}, extreme strength-to-weight ratios\,\citep{Qin2017_gyroid_graphene}, and simultaneous stiffness and stretchability\,\citep{Surjadi2025_DoubleNetwork}. These features enable applications ranging from thermal management\,\citep{Fan2022_BatteryThermalManagement,ATTARZADEH2022_TPMSHeatExchangers} to biomedical implants\,\citep{Arash2018_BoneImplants,Ambu2019_BoneImplants}. Yet the design space is effectively limitless, and its full potential remains unexplored.

Metamaterial discovery typically follows two paradigms: \emph{forward design}, which proposes a structure and then measures its properties, and \emph{inverse design}, which begins with target properties and then searches for a matching structure. Both workflows demand (i) domain expertise, (ii) a grasp of relevant material metrics, (iii) concise yet expressive geometric representations, and (iv) algorithms that map between structural and functional spaces.

Vision–language models (VLMs) are well suited to this challenge, as they excel at the cross-modal reasoning, retrieval, and generation required for effective metamaterial design -- spanning text, images, 3-D geometry, and numerical property vectors. 
The complex, verifiable data given by metamaterial design tasks also offers an ideal sandbox for VLM and AI research targeting real-world applications.

Despite this symbiotic potential, data-driven metamaterial design is hindered by several issues. 
For example, \citet{Surjadi2025_PerspectiveSurvey} cite the need for “universal tools capable of parametrizing varied architected material morphologies." 
There is also an acute need for reusable, reconfigurable, task-agnostic datasets featuring diverse structure architectures \citep{Lee2024_DataDrivenMetamaterialsSurvey}.

To address these gaps and catalyse progress in both communities, we introduce the first foundational VLM ecosystem for metamaterial design, anchored by three components:

\begin{enumerate}
    \item \textbf{MetaDSL}: a domain-specific language that captures metamaterials in a structured, compact, and expressive form accessible to both humans and large language models.
    \item \textbf{MetaDB}: a database of more than \numMaterialsInDatabaseRoundedDown{} metamaterials, each of which pairs a MetaDSL program with the derived 3-D geometry, rendered images, and simulated properties.
    \item \textbf{MetaBench}: benchmark suites that probe three fundamental metamaterial design tasks -- structure reconstruction, property-driven inverse design, and performance prediction -- using data sampled from MetaDB.
    %\item \textbf{MetaAssist}: a VLM assistant fine-tuned on MetaBench and deployed within an interactive, CAD-like environment that facilitates multi-modal design interactions including language, images, geometry, and MetaDSL code.
\end{enumerate}

To complete our vision, we also use MetaBench to train and evaluate \textit{MetaAssist}, a VLM assistant baseline and interactive CAD environment that facilitates multi-modal design interactions including language, images, geometry, and MetaDSL code. 

All four components are designed for extensibility and community contribution, such that they can evolve seamlessly alongside the state of the art in materials science and agentic design. 
Collectively, our ecosystem provides a coherent, extensible knowledge base for metamaterial design, while laying the foundation for intuitive, efficient human–AI collaboration in architected materials.%this domain.

\section{Background}
\label{sec:background}

\paragraph{Metamaterials}
Two long-standing hurdles in metamaterial design are (i) navigating the immense geometric diversity of candidate architectures and (ii) modelling the intricate, often non-linear mapping from geometry to effective properties\,\citep{Makatura2023_ProcMeta,Lee2024_DataDrivenMetamaterialsSurvey,xue2025_mindmicrostructureinversedesign, Surjadi2025_PerspectiveSurvey}.  Many studies rely on trial-and-error \emph{forward} design, where experts hand-craft parameterized structures for specific targets\,\citep{Jennet_DiscreteMetamaterials_2020,Muhammad2021_3dPhononicBandGap, Frenzel2017_Twist, Meier2025_bandgap}. Now, data-driven pipelines also provide a more scalable, systematic alternative in the service of \emph{inverse} design: \citet{Panetta2015_ElasticTextures} analysed $1205$ families of cubic truss lattices, while \citet{AbuMualla2024_SymmetryInducedDesign} expanded to $17\,000$ truss structures spanning six crystal lattices.  Beyond trusses, high-throughput workflows create thousands of thin-shell architectures including plate lattices\,\citep{Sun2023_PPL} and TPMS-inspired surfaces\,\citep{Xu2023_NewTPMSFamilies,Liu2022_PSL,Yang2022_HighThroughput,Chan2020_METASET}.  Because many datasets target a single architecture class (\eg beams or shells) and a narrow performance metric, they restrict the attainable property gamut and thus the capability of downstream models\,\citep{Berger2017_StiffnessLimit,Lee2024_DataDrivenMetamaterialsSurvey}.  
Recent designs also increasingly blend classes in hybrid or hierarchical forms\,\citep{Surjadi2025_DoubleNetwork, Chen2019_HybridTPMS, White2021_IPL}, emphasising the need for representations that span such boundaries.
The procedural-graph approach of \citet{Makatura2023_ProcMeta} captures diverse geometries but is demonstrated primarily for human-in-the-loop workflows.  Voxel and hybrid encodings scale to 140\,k–180\,k diverse structures\,\citep{yang2024guideddiffusionfastinverse,xue2025_mindmicrostructureinversedesign}, but they sacrifice semantic clarity and compactness, which complicates human or agent editing. Such tradeoffs -- along with inconsistencies in geometry descriptors, vocabularies, and evaluation protocols -- continue to impede dataset reuse and extensibility\,\citep{Lee2024_DataDrivenMetamaterialsSurvey}.  

We close these gaps with a universal metamaterial descriptor (MetaDSL) along with a reconfigurable database of \numMaterialsInDatabaseRoundedDown{} metamaterials (MetaDB). Each MetaDB entry couples a succinct, semantically rich program with derived 3-D geometry, renderings, and simulated properties, enabling consistent comparison and seamless expansion.  Programmatic templating further enlarges the design space, and community contributions can grow both MetaDB and the accompanying benchmark suite.

\paragraph{Vision–Language Models for Design}
Large language and vision–language models (VLMs) have recently permeated design tasks, including procedural textures\,\citep{li2025vlmaterial}, 3-D scenes\,\citep{yang2024holodeck,kumaran2023scenecraft}, mesh generation and editing\,\citep{sun20233d,wang2024llamameshunifying3dmesh,jones2025shapelib,huang2024blenderalchemy,yamada2024l3go}, interior layouts\,\citep{celen2024idesign}, sewing-pattern synthesis\,\citep{nakayama2024aipparel,bian2024chatgarment}, and computer-aided engineering and manufacturing \citep{Makatura2024a_LLMsForCDAM_pt1,Makatura2024b_LLMsForCDAM_pt2, Choi2025_GenPara, Yuan2024_CADTalk}. In most cases, code serves as the medium: pretrained models follow instructions, reuse standard patterns, and emit domain-specific scripts (\eg Blender Python).  When tasks demand novel grammars or specialist knowledge, fine-tuning further elevates performance\,\citep{zhou2024design2garmentcode}.

Our work adopts this code-centric philosophy but tailors it to metamaterials, whose design demands rich geometric semantics, strict physical constraints, and fluid translation among text, images, parameterised programs, and numerical property vectors.  By grounding the interface in a purpose-built DSL and a physically validated database, we lay a robust foundation for future VLMs to reason about, generate, and refine architected materials at scale.

% ====================================
%       Language
% ====================================
\section{Domain-Specific Language}
\label{sec:dsl}

\begin{figure}[t]
    \centering
    \includegraphics[width=\linewidth]{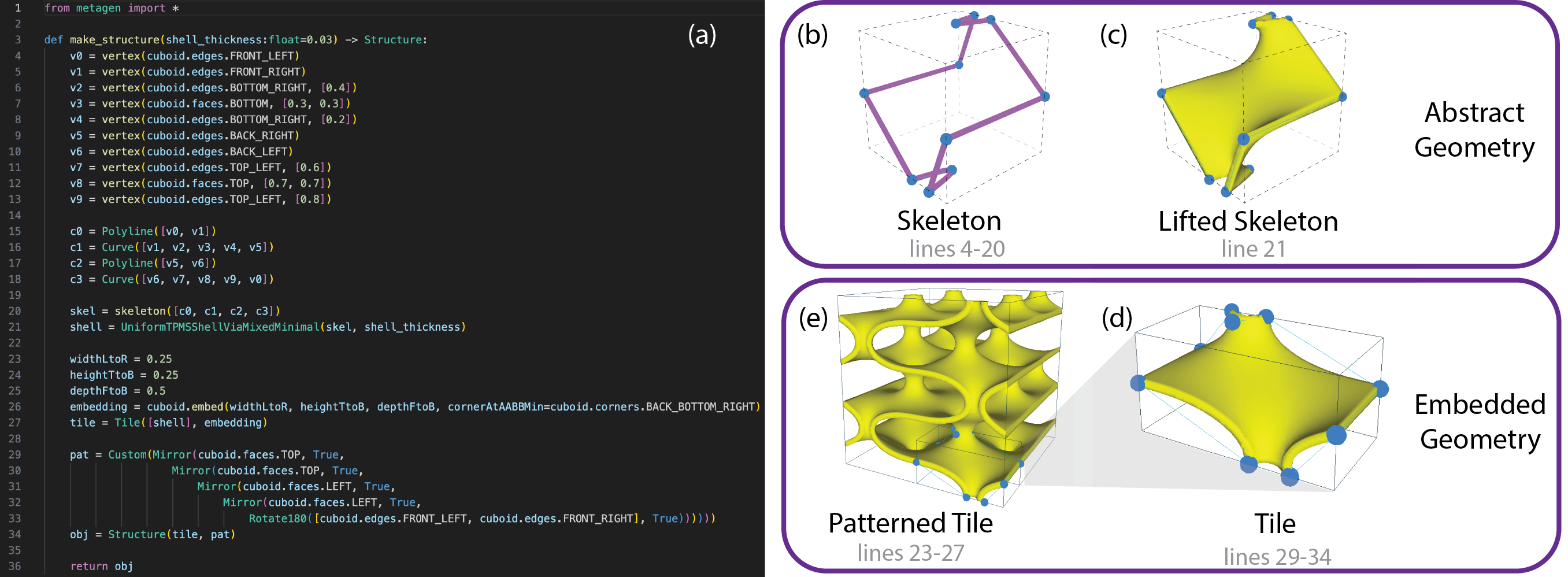}
    \caption{A MetaDSL program (a) and illustrations of each construction stage (clockwise): \textbf{(b)} build a 1D skeleton relative to an abstract convex polytope CP -- here, a cuboid; \textbf{(c)} specify a lifting procedure from 1D to 3D; \textbf{(d)} embed the CP in $\R^3$ to create a tile, and execute the lifting procedure to create our final geometry; and finally, \textbf{(e)} tessellate the tile according to the specified pattern.}
    \label{fig:dsl_stages}
\end{figure}

\begin{figure*}[t]
    \centering
    \includegraphics[width=\linewidth]{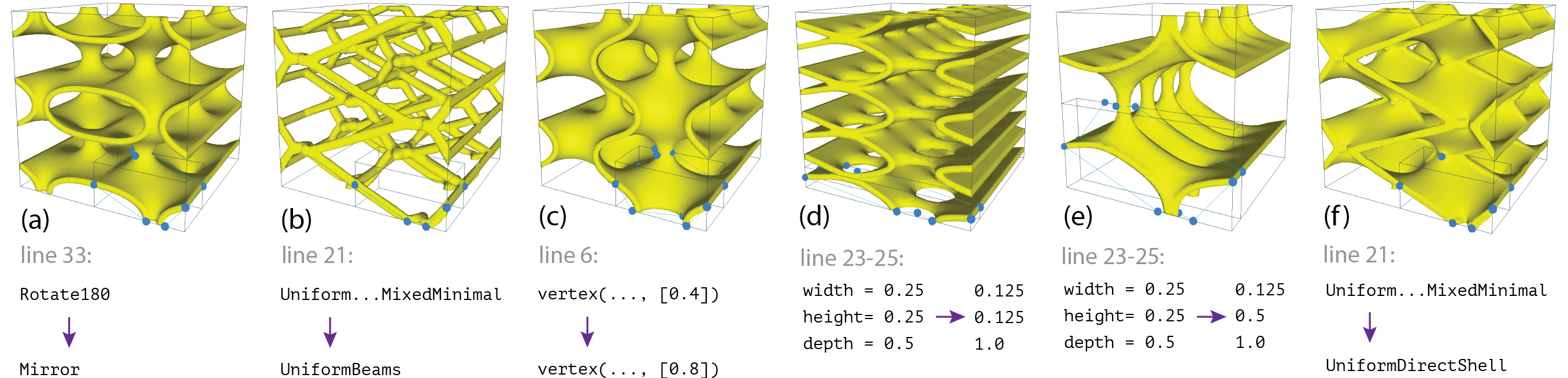}
    \caption{We illustrate the expressive power of MetaDSL by showing six different structures that all stem from the program shown in \Cref{fig:dsl_stages}(a). Each one is produced by changing a single aspect of the original program, as detailed below each structure.}
    \label{fig:seed_variants}
\end{figure*}

To support our vision of an expansive, dynamic metamaterial ecosystem, a suitable structure representation is key. 
An ideal representation would 
(1) support the full range of metamaterial architectures; 
(2) facilitate modularity and reuse;
(3) be compact, semantically meaningful, and easy to use;
(4) be amenable to and robust under generative design;
(5) encourage valid metamaterials by construction;
and
(6) be quickly verifiable through type-checking.
In designing MetaDSL, we laid out a long-term design philosophy that is amenable to all of these goals.
Although our current implementation realizes a core subset of this functionality (detailed in \Cref{sec:dsl-implementation}), the infrastructure is built with extensibility in mind.
This will facilitate the continued development of MetaDSL, such that new design paradigms can be added to MetaDSL as the field matures, without invalidating existing programs.

\subsection{Language Design Philosophy}
Using MetaDSL, materials are defined by a combination of modular, reusable components. 
A rich type system determines the compatibility between components at different levels, which allows for programmatic composition with verifiable outcomes.

Broadly, these components follow a bi-level approach that is common for metamaterial design.
The first level describes a small representative unit of the structure, called a \textit{tile}.
The second level specifies a \textit{pattern}: transformations that extend a tile into a space-filling structure. 
In MetaDSL, these layers are independent and polymorphic: a pattern can be applied to any number of tiles, and vice-versa.

A detailed view of the MetaDSL construction process is shown in \Cref{fig:dsl_stages}.
The first stage specifies a \emph{skeleton}, which is a set of open or closed 1D curves defined relative to an abstract convex polytope (CP)-such as a cuboid or a tetrahedron. To facilitate downstream compatibility checks, our type system classifies the skeleton by examining its topology and its relationship to the CP boundaries.  
Next, we select a \emph{lifting function} that will be used to promote 1D skeleton curves into 3D geometry. The applicability of a lifting function to a skeleton is determined by its type.
The third stage specifies a concrete embedding in $\R^3$ for our CP. At this stage, our lifted skeleton can be evaluated to yield the final structure geometry and thus, a completed \emph{tile}.
This separation between the abstract CP of a skeleton and the concrete CP of a tile is subtle but critical, as it permits compositional re-use of skeletons, as demonstrated in \Cref{fig:seed_variants}(d,e). 
However, it is also essential that we assign an embedding before proceeding, because the admissible pattern operations are influenced by extrinsic geometric measures such as the dihedral angles between the polytope planes.
To promote the tile into a space-filling object, MetaDSL applies a \emph{pattern} composed of spatial repetition procedures like mirroring, gliding, rotating, \etc. Patterns themselves can be also be composed into larger patterns.
In a final layer, we also provide standard constructive solid geometry (CSG) Boolean operations to combine multiple structures.
This makes it easy to define structures with mixed scales, multiple symmetry classes, or interpenetrating lattices \citep{White2021_IPL}.

This philosophy supports the stated goals for our representation in myriad ways.
For example, because vertex positions are specified relative to their parent CP (\eg, at the halfway point of \dslFrag{cuboid.edges.TOP\_LEFT}), it is easy to identify valid position bounds; this facilitates robust exploration. Inclusion of common synonyms in the syntax hardens against common LLM hallucinations (e.g. \dslFrag{TOP\_LEFT} and \dslFrag{LEFT\_TOP}).
The separation between abstract skeletons, embedded tile geometries, and patterns encourages modular re-use at multiple scales.
Moreover, deriving tile geometry from abstract skeletons enables pattern compatibility verification based on boundary adjacency, and the library of lifting functions covers metamaterial design patterns in existing literature.

\subsection{Implementation Details}
\label{sec:dsl-implementation}

We implement our language as an embedded DSL in Python, which provides a familiar, flexible interface with support for comments, descriptive identifiers, higher-order templates, and parameterization.
We use Procedural Metamaterials (\procMeta) \citep{Makatura2023_ProcMeta} as our geometry kernel, as their representation is specifically designed to capture a variety of metamaterial classes.
Targeting this backend introduced limitations which influenced the core functionality we implemented.
For example, \procMeta only supports materials with translational-units that reside in a unit cube; thus, MetaDSL currently lacks support for patterns beyond that scope.
The \procMeta skeletal design space also directly informed the abstract CPs and lifting functions that we implemented in MetaDSL.
However, as MetaDSL can be transpiled to any kernel, it is not inherently bound by these limitations. If a more general backend were developed, our language could be extended to accommodate the larger feature set without invalidating our existing examples. \Cref{supp:MetaDSL} gives a detailed description of the language design, implementation, system design insights, and comparison to \procMeta; \Cref{sec:app-metadsl-api} contains the complete MetaDSL documentation.

%% File to use
% ====================================
%       Dataset Section
% ====================================
\section{Database Generation}
\begin{figure*}[t]
    \centering
    \includegraphics[width=\linewidth]{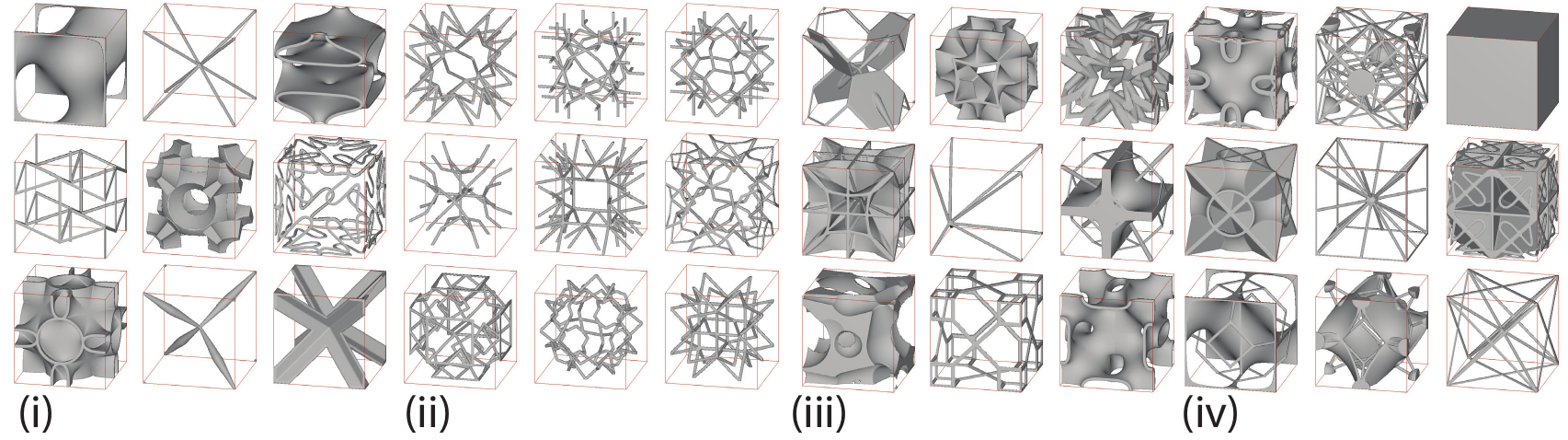}
    \caption{Assortment of metamaterials in MetaDB, illustrating four creation modes: (i) hand-authored seeds, (ii) generated models, (iii) type-enabled mutations, and (iv) LLM-augmented hybrids.}
    \label{fig:metadb-assortment}
\end{figure*}

MetaDSL represents metamaterials in a consistent, concise manner, which permits a single pipeline that produces code, watertight geometry, renderings, and simulated properties for \emph{every} entry. 
% We first explain how models are created, including augmentation strategies that diversify the corpus. 
To ensure the quality of MetaDB, we only add \textit{validated} models that pass basic checks (see \Cref{supp:metadb-quality}).

\subsection{Constructing Metamaterial Models}

Each metamaterial is a DSL program, or \emph{model}, that may optionally expose a set of design parameters (with default values).
Our metadata block also allows program authors to include details such as bounds, dependencies, or recommended ranges for each parameter. This clarifies design freedom, enables continuous exploration, and provides hooks for optimisation schemes. The metadata is stored in a machine-parsable format (YAML) with a prepopulated scheme for tracking \eg provenance, versioning, and notable traits about the structure, including symmetries, architecture type (beams, shells, \etc), and related structures.
Our metadata also permits custom fields.

\paragraph{Direct Construction}
\emph{Authored} models are human-written, with provenance records tracking the model author and the original design source(s), and editable semantic parameters to encode families of models.  We also provide a programmatic \emph{generator} interface to create families of models.  As a proof of concept, we implemented a generator following \citet{Panetta2015_ElasticTextures}; this generates parametrized models for all 1,205 truss topologies using a few hundred lines of Python. Our type-checked DSL allows us to specify and evaluate validity constraints on the small tile, without needing to generate the fully-patterned beam network. Moreover, because our generator is exposed and editable, we can easily modify the high-level generator parameters (\eg maximum vertex valence) to output different sub- or supersets of interest. For each \emph{generated} model, the provenance metadata stores the generator script, its settings, and per-instance parameters; generator parameters may be substituted for specific values or passed through to remain exposed in the resulting programs.

\paragraph{Augmentation}
We propose two orthogonal protocols to enlarge MetaDB based on existing models. 
Our first strategy, \emph{Hybridization} (crossover), is motivated by works that offer unique, extremal mechanical properties by hybridizing common structures such as trusses+woven beams \citep{Surjadi2025_DoubleNetwork}, nested trusses \citep{Boda2025_NestedTrusses}, TPMS shells+planar shells \citep{Chen2019_HybridTPMS}, and trusses+solids \citep{White2021_IPL}. 
We emulate this process by prompting an LLM with pairs or triplets of parent programs, then requesting hybrid code.
Our prompting strategy (detailed in \Cref{supp:MetaDBHybridization}) follows insights from recent works in LLM-mediated program search \citep{li2025vlmaterial, RomeraParedes2024_FunSearch}.
The resulting \emph{hybridized} model stores its parent IDs, prompt details, and LLM details as provenance information. 

Our second strategy, \emph{mutation}, leverages MetaDSL's type system to apply targeted edits—such as skeleton reconfiguration, pattern adjustment, and lift procedure changes—while guaranteeing validity. 
The operators are described in \Cref{supp:MetaDBMutation}.
These operations are motivated by works such as \citet{Akbari2022_StrutTPMS}, which posits beam approximations of TPMS shells.
Each mutation stores its parent and details about the mutator function.

\subsection{Auxiliary Data Generation}
\label{sec:database-aux-data-gen}

For every model we generate three auxiliary artifacts: geometry, renderings, and physical property predictions.
To obtain the geometry, we transpile our MetaDSL model into a \procMeta{} graph\,\citep{Makatura2023_ProcMeta} and use their geometry kernel to export a watertight \texttt{.obj}.
Using the exported mesh, our custom \textsc{PyRender} scene produces orthographic images from the front, top, right, and front-top-right viewpoints.
Finally, we use the integrated simulations of \procMeta{} to voxelize the mesh on a $100^{3}$ grid and perform periodic homogenisation using a base material with $E{=}1$, $\rho{=}1$, $\nu{=}0.45$.  
The resulting $6{\times}6$ stiffness matrix $C$ is reduced to 18 scalars: six global metrics—Young’s modulus $E$, shear modulus $G$, Poisson ratio $\nu$, bulk modulus $K$, anisotropy $A$, volume fraction $V$—plus directional values for $E$ (3), $G$ (3), and $\nu$ (6).  More details are available in \Cref{sec:app-database-properties}.
MetaDB therefore combines code, geometry, simulation, imagery, and rich provenance—providing a unified benchmark and a data-efficient training ground for vision–language metamaterial assistants.

%% File to use!

% ====================================
%       Task Generation & Benchmark
% ====================================
\section{Benchmark Curation}
\label{sec:benchmark}

From MetaDB we derive a benchmark that covers three fundamental
metamaterial tasks:
\textbf{(1) reconstruction}—produce a DSL program that reproduces a target
structure (for example, from images);
\textbf{(2) material understanding}—predict the property profile of a given
structure description; and
\textbf{(3) inverse design}—generate a DSL program that satisfies a
requested property profile.
Each task supports multiple \emph{query types} based on the inputs available. For instance, material understanding may be invoked with a single image (``1-view'') or with four images plus code (``multiview\_and\_code''). The benchmark suite ships a dataset for every
query type.

% ------------------------------------------------------------------
\subsection{Task-Based Dataset Construction}
We start with a designated pool of \emph{active} models and partition them into train, validation, and test splits that remain fixed for all tasks. The relevant information for each query type is as follows.

\paragraph{Reconstruction.}
Given $n\!\in\!\{1,\dots,4\}$ orthographic images, the desired output is a DSL program whose rendered geometry matches the target. Because every model has four views (Section~\ref{sec:database-aux-data-gen}), each model contributes $\binom{4}{n}$ examples to the $n$-view dataset.

\paragraph{Material understanding.}
Given a structure description, the desired output predicts six global properties: Young’s
modulus $E$, shear modulus $G$, bulk modulus $K$, Poisson ratio $\nu$,
anisotropy $A$, and volume fraction $V$. Values are rounded to two
significant figures. Our benchmark supports two query types:
\textit{multiview\_and\_code} (four images + DSL code) and
\textit{single\_image} (one image). The relative performance on each type indicates whether additional context helps or hinders a given VLM.

\paragraph{Inverse design.}
Given a target property profile, the desired output is a DSL program whose simulated properties satisfy the profile. We generate datasets for six query types, where the length-$n$ query requests  $n\!\in\!\{1,\dots,6\}$ property targets per profile. Targets may be exact values,
ranges, or upper/lower bounds—\eg, ``auxetic ($\nu<0$)'' or
``volume fraction $V\approx0.6$.'' To construct target profiles from a model, we (1) sample $n$ active properties from the model, (2) choose bounds for each, and (3) render a natural-language prompt using a grammar conditioned on each property’s part-of-speech tag (adjective, verb, etc.).
This process is detailed in \Cref{sec:app-dataset-inv-design-prompt-construction}.
Both the prompt and the underlying numeric targets are stored, so users can
rephrase questions or bypass NLP entirely.

\paragraph{Omnitask dataset.}
For completeness, we provide an \emph{omnitask} split that unites every query type into a
single corpus; this is useful for training generalist agents.

\subsection{Task-Based Example Format}
The query/response pairs are constructed using prompt templates that are specific to each task type (listed in \Cref{sec:querytemplates}). 
Given a metamaterial and a task type, we first gather the data that will be used to construct the query/ground truth response, along with the information required to evaluate the predicted response. The intermediate format used to organize this information is detailed in \Cref{sec:app-dataset-intermediate-rep}. In addition to being model agnostic, this intermediate format allows researchers to reframe prompts without regenerating or deviating from the core content of the inquiry. The intermediate representation also makes MetaBench applicable to traditional non-AI methods. However, since no traditional methods cover the full breadth of MetaBench, we do not include traditional baselines in our evaluations.

\ignore{
}

\ignore{

}

\ignore{
% internally tracks incidence on the CP, structure of the final skeleton to judge validity for subsequent input steps, with plain english descriptions about why the input is invalid -- a burden typically left to the human/AI author in previous works. eg, in the user study supplement of \citet{Makatura2023_ProcMeta}, they talk about having to intervene if the user was about to try generating a structure that would crash the program due to invalid input.
}
\section{Results} % This could be called "Experiments" or similar
\label{sec:results}

\subsection{Database}
MetaDB is, to our knowledge, one of the largest metamaterial databases ever collected, comprising \numMaterialsInDatabase materials. Our dataset features 36,997 expert material designs, including 1,588 variations of 50 hand-authored programs, 1,205 generations, and 34,204 generation parameter variations. We also introduce 12,029 hybrids and 141,234 mutations. 

To validate MetaDB, we examine its property gamut relative to our expert seeds. The property gamut of MetaDB compares favorably to that of our expert seed programs. 
Both are centered around similar ranges, suggesting that our design space is valid and relevant. 
However, MetaDB offers more uniform, dense coverage, along with a wider range for properties like anisotropy (\textasciitilde 2x the expert range) and directional Poisson ratios (\textasciitilde 1.2-4x the expert range).

%the property range is x
%our augmentation schemes are interesting, generate direct hybrids and also non-trivial combinations -- eg finding loops in a beam structure and instantiating a surface over that instead

\subsection{Benchmark \& Baseline}

%We'll compare the ground truth (gt) to the predicted results (pred).

\begin{table}[]
    \centering
        \begin{tabular}{cccccccc}
            \toprule
            Category & \multicolumn{2}{c}{Inverse Design} & \multicolumn{2}{c}{Material Understanding} & \multicolumn{3}{c}{Reconstruction} \\
            Metric & Error & Valid & Error & Valid & CD & IoU & Valid \\
            Model &  &  &  &  &  &  &  \\
            \midrule
            LLaVAOmniTask & \textbf{0.011} & \textbf{91.9\%} & 0.024 & \textbf{100\%} & 0.034 & 0.490 & 82.9\% \\
            LLaVASingleTask & 0.036 & 81.9\% & \textbf{0.018} & \textbf{100\%} & \textbf{0.029} & \textbf{0.524} & 83.8\% \\
            NovaLite & 0.060 & 2.7\% & 0.200 & \textbf{100\%} & 0.119 & 0.051 & 19.3\% \\
            NovaOmniTask & 0.026 & 91.4\% & 0.032 & \textbf{100\%} & 0.045 & 0.334 & \textbf{87.2\%} \\
            NovaSingleTask & 0.032 & 79.2\% & 0.153 & \textbf{100\%} & 0.059 & 0.205 & 84.8\% \\
            OpenAIO3 & 0.038 & 24.7\% & 0.077 & \textbf{100\%} & 0.053 & 0.147 & 54.6\% \\
            \bottomrule
        \end{tabular}
    \caption{Category-level evaluation results for various models on MetaBench. Average Normalized Error (Error), Chamfer Distance (CD), and Intersection over Union (IoU) are averaged over valid responses for across all tasks within a category, and Valid reports the percentage of valid responses (note that this means SingleTask models are averaging over fewer task examples). Responses are considered invalid if they do not contain code or the requested prediction metrics, or if the generated program does not produce a valid metamaterial. Untuned LLaVa-Next values are not reported because it failed to produce any valid outputs. See \Cref{fig:reconstructiongallery,fig:inversedesigngallery,fig:understandinggallery} for qualitative evaluation.}
    \label{tab:benchmarksummary}
\end{table}

The 13,282 authored, generated, and hybrid models form the core set from which MetaBench is sampled. We randomly split these models into 500 test, 50 validation, and 12,732 training materials, and generated benchmark tasks for each as described in \Cref{sec:benchmark}.

We tested a variety of commercial and open source VLMs on MetaBench, both fine-tuned and zero-shot. \Cref{tab:benchmarksummary} summarizes these models' performance at the task category level; additional tables and galleries in \Cref{supp:extrabench} break down performance at the task level, with confidence intervals and qualitative interpretations. These experiments revealed three primary insights. Firstly, that fine-tuning is necessary for strong performance on MetaBench. Untuned models are generally unable to produce consistently valid programs, though when they do, a reasoning model (o3) can perform in-line with weaker tuned models. Secondly, fine-tuning generalist multi-task models improves inverse design performance. Finally, a tuned small model outperforms a tuned large model in nearly all metrics. However, this is likely due to it being able to converge more quickly given the same training budget (see \Cref{supp:trainingcurves}). Error metric definitions and tested model details are given below.

\paragraph{Material Reconstruction}
Reconstruction measures 3D structure similarity, measured by intersection over union (IoU) and volumetric chamfer distance of the voxelized unit cells.
%We measure performance using two metrics based on the 3D geometry generated by the structure.
%First, we report the Intersection Over Union (IOU), which measures the relative overlap between a pair of voxelized structures -- in this case, $\frac{\textrm{vox}_{\textrm{pred}} \cap \textrm{vox}_{\textrm{gt}}}{\textrm{vox}_{\textrm{pred}} \cup \textrm{vox}_{\textrm{gt}}}$. 
%We also measure the chamfer distance between \tofill{voxels? intuitive description}.
%As evident in the table/figure \tofill{discuss results}

\paragraph{Material Understanding}
Material understanding is computed as an Averaged Normalized Error across six properties: anisotropy, Young's modulus (VRH), Bulk Modulus (VRH), Shear Modulus (VRH), Poisson's Ratio, and Volume Fraction, normalized to the typical range of that property across the core material set.%The absolute prediction error for each property is divided by a property dependent normalizing factor that is the typical range of that property across the core material set.

\paragraph{Inverse Design}
Inverse design is measured by a clipped Averaged Normalized Error. For specific value targets normalized error is computed as above. For bounds targets, normalized error is taken relative to the bound (and is zero if the bound is respected).

\paragraph{Models}
We tested 3 base models; a small open-source VLM (LLaVA-Next), a large commercial VLM (Amazon Nova Lite), and a large commercial chain-of-thought reasoning model (Open AI o3 with medium reasoning). For LLaVA-Next and Nova Lite we also produced four fine-tuned variants trained on the MetaBench training set. The OmniTask variants were tuned over all training task examples in MetaBench, and three SingleTask variants were trained over one category-representative task type each (4-view reconstruction, 4-target inverse design, and multi-view-plus-code material understanding). \Cref{tab:benchmarksummary} condenses the SingleTask variants of each model into a single row for compactness. In initial tests, untuned LLaVA-Next failed to produce any valid output, likely due to the MetaDSL description overwhelming its context window, so has been excluded from the table.

\paragraph{Implementation}
LLaVA models are tuned from Llama3-LLaVA-Next-8b~\cite{li2024llava, liu2024llavanext}. Commercial models were tuned and tested with default settings. All tuned models were trained for 1 epoch. For full training and inference details, see \Cref{supp:implementation}. Benchmark construction and model prompts are detailed in \Cref{supp:MetaBench} and \Cref{sec:querytemplates}.

\subsection{Interactive Case Studies}

We built a browser-based metamaterial copilot interface to explore practical scenarios. It consists of a VLM chat window on the left, a code editor in the center, and a material preview window on the right. We conducted a series of case studies, using NovaLiteOmniTask as our interactive model due to its large context window and stronger conversational abilities. We experimented with a variety of prompts, and present here two scenarios that illustrate the potential of a metamaterial design copilot.

The first is creating a material from an input image. Images are compelling input for material design because they cover trying a new material described pictorially in literature, sketching an idea for a design, or taking inspiration from a structure in nature. We prototyped this functionality with a material from the MetaBench test set; even though we presented our request conversationally rather than in structured form, we were still able to obtain and fabricate a perfect reconstruction.

\begin{figure}
    \centering
    \includegraphics[width=\linewidth]{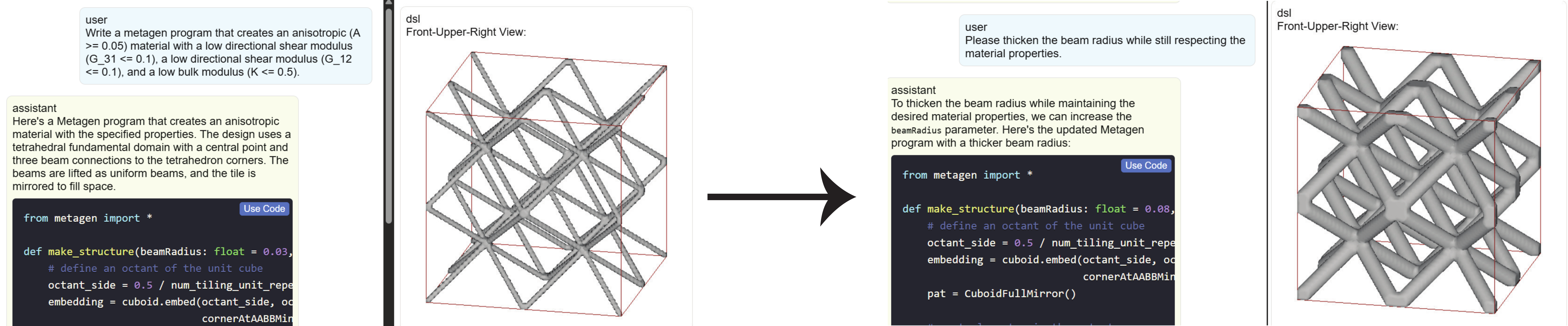}
    \caption{Iterative Inverse Design: Designers can specify desired target properties, and these preferences and constraints can be considered throughout multiple design iterations.}
    \label{fig:inversedesignstudy}
\end{figure}

The second is iterative inverse design. In \Cref{fig:inversedesignstudy}, we specify a set of target property bounds, and the model is able to generate a metamaterial that satisfies them (we verified this with our simulator). But design is always iterative, and seeing one design can spark new criteria and objectives. In this case we wanted a thicker structure that still conformed to our original input, and (again verified by simulation), the model was able to update the design within target parameters. This illustrates the powerful ability of language models to remember and carry through \emph{design context}, allowing for assistance across multiple design iterations.

\begin{figure}
    \centering
    \includegraphics[width=\linewidth]{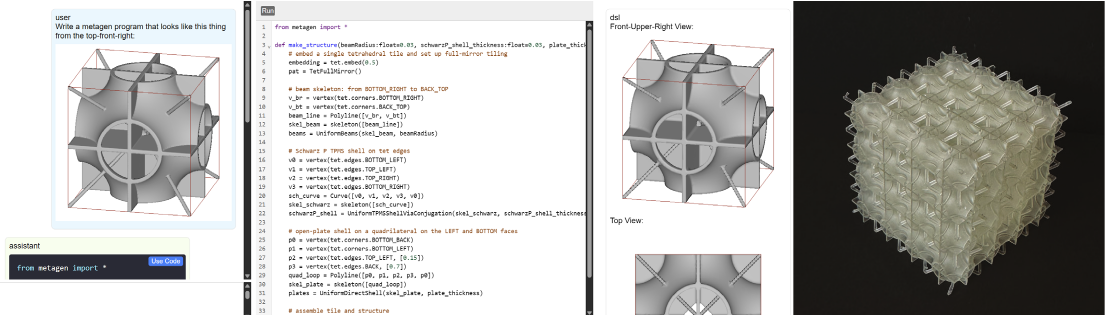}
    \caption{Reconstruction: (Left) Generating a metamaterial program from an input image enables incorporating designs from literature, sketches, and nature. (Right) 3D printed design.}
    \label{fig:reconstructionstudy}
\end{figure}

\ignore{
\liane{moved from the old methods/DSL intro -- leaving here in case we want this or something similar for the comparisons}
%Our early experiments with the representation of \citet{Makatura2023_ProcMeta}(``\procMeta'') quickly revealed that the VLM requirements went beyond those that had been considered for efficacy in general computational methods. 
% Specifically, we found that the VLM struggled with \procMeta's continuous spatial parameter requirements, such as the absolute position of each vertex in $R^3$. 
% The VLM also had difficulty understanding the implicit hierarchical dependencies and constraints that exist among potentially distant components in the \procMeta graph representation.
% Finally, the VLM struggled with a lack of semantically meaningful content in the graph specification, which was given in a custom JSON format.
% 

We compare extensively to the approach of \citet{Makatura2023_ProcMeta}, as it is both the inspiration and the closest available baseline for our DSL. 

\subsection{Experiments}
We evaluated and ablated metagen over four axes of variation; VLM model, data augmentation representation, data representation, and query strategy. Models were Nova-Lite, LlaVa, and GPT-4o; for Nova-Lite and LlaVa we evaluated both tine-tuned (-FT) and base models. Data and Data Augmentation representations are either Graph or DSL, where Graph means the VLM was shown an intermediate graph representation (procmeta) rather than the python DSL. The Augentation Representation is the representation used to hybridize and create the training data set, while the query representation is the format the VLM is asked to generate in. Because DSL programs map to Graphs, but not visa-versa, graph augmentations are only usable with graph queries, and evaluate the effect of representation on date generation. We tried 3 querying strategies; 0-shot prompting where the model is prompted with a description of the material representation and the generation requirements, and two few-shot in-context learning (ICL) strategies, one where K random example programs are given (along with their properties and images), and one where the K nearest training programs \tofill{(TODO - how to measure closeness with multiple non-normalized metrics?)} are given as examples. \Cref{tab:generationresults} shows the results. \tofill{(TODO - analysis)} \todo{reword nicely}

\begin{table}
\caption{Valid Material Generation Rate}
\label{tab:generationsuccessrates}
\begin{tabular}{lllr}
\toprule
 &  &  & Success Rate \\
Input & Context & Model &  \\
\midrule
\multirow[t]{11}{*}{Images} & \multirow[t]{3}{*}{5-NN} & NovaLite & 0.20 \\
 &  & NovaPro & 0.38 \\
 &  & NovaProFinetuned & 0.61 \\
\cline{2-4}
 & \multirow[t]{2}{*}{5-RN} & NovaLite & 0.13 \\
 &  & NovaPro & 0.27 \\
\cline{2-4}
 & \multirow[t]{6}{*}{0-Shot} & LLaVA & 0.01 \\
 &  & LLaVA-Finetuned & 0.95 \\
 &  & NovaLite & 0.04 \\
 &  & NovaLiteFinetuned & 0.80 \\
 &  & NovaPro & 0.00 \\
 &  & NovaProFinetuned & 0.84 \\
\cline{1-4} \cline{2-4}
\multirow[t]{9}{*}{Images+Properties} & \multirow[t]{3}{*}{5-NN} & NovaLite & 0.19 \\
 &  & NovaPro & 0.41 \\
 &  & NovaProFinetuned & 0.59 \\
\cline{2-4}
 & \multirow[t]{2}{*}{5-RN} & NovaLite & 0.16 \\
 &  & NovaPro & 0.32 \\
\cline{2-4}
 & \multirow[t]{4}{*}{0-Shot} & NovaLite & 0.02 \\
 &  & NovaLiteFinetuned & 0.78 \\
 &  & NovaPro & 0.00 \\
 &  & NovaProFinetuned & 0.84 \\
\cline{1-4} \cline{2-4}
\multirow[t]{8}{*}{Properties} & \multirow[t]{2}{*}{5-NN} & NovaLite & 0.41 \\
 &  & NovaPro & 0.51 \\
\cline{2-4}
 & \multirow[t]{2}{*}{5-RN} & NovaLite & 0.30 \\
 &  & NovaPro & 0.59 \\
\cline{2-4}
 & \multirow[t]{4}{*}{0-Shot} & NovaLite & 0.01 \\
 &  & NovaLiteFinetuned & 0.87 \\
 &  & NovaPro & 0.02 \\
 &  & NovaProFinetuned & 0.85 \\
\cline{1-4} \cline{2-4}
\bottomrule
\end{tabular}
\end{table}

\begin{table}
\caption{Reconstruction Accuracy from Images}
\label{tab:imagereconstruction}
\begin{tabular}{lllrrr}
\toprule
 &  &  & IoU & CD & IoU (all) \\
Props? & Context & Model &  &  &  \\
\midrule
\multirow[t]{11}{*}{No} & \multirow[t]{3}{*}{5-NN} & NovaLite & 0.235 & 0.008 & 0.046 \\
 &  & NovaPro & 0.220 & 0.009 & 0.083 \\
 &  & NovaProFinetuned & 0.206 & 0.010 & 0.126 \\
\cline{2-6}
 & \multirow[t]{2}{*}{5-RN} & NovaLite & 0.134 & 0.012 & 0.018 \\
 &  & NovaPro & 0.152 & 0.009 & 0.042 \\
\cline{2-6}
 & \multirow[t]{6}{*}{0-Shot} & LLaVA & 0.008 & 0.011 & 0.000 \\
 &  & LLaVA-Finetuned & 0.465 & 0.003 & 0.443 \\
 &  & NovaLite & 0.077 & 0.014 & 0.003 \\
 &  & NovaLiteFinetuned & 0.252 & 0.007 & 0.202 \\
 &  & NovaPro & N/A & N/A & 0.000 \\
 &  & NovaProFinetuned & 0.244 & 0.007 & 0.206 \\
\cline{1-6} \cline{2-6}
\multirow[t]{9}{*}{Yes} & \multirow[t]{3}{*}{5-NN} & NovaLite & 0.207 & 0.008 & 0.040 \\
 &  & NovaPro & 0.286 & 0.008 & 0.119 \\
 &  & NovaProFinetuned & 0.257 & 0.009 & 0.151 \\
\cline{2-6}
 & \multirow[t]{2}{*}{5-RN} & NovaLite & 0.109 & 0.010 & 0.017 \\
 &  & NovaPro & 0.166 & 0.009 & 0.053 \\
\cline{2-6}
 & \multirow[t]{4}{*}{0-Shot} & NovaLite & 0.046 & 0.012 & 0.001 \\
 &  & NovaLiteFinetuned & 0.250 & 0.008 & 0.196 \\
 &  & NovaPro & N/A & N/A & 0.000 \\
 &  & NovaProFinetuned & 0.287 & 0.007 & 0.241 \\
\cline{1-6} \cline{2-6}
\bottomrule
\end{tabular}
\end{table}

\begin{table}
\caption{Inverse Design for Elasticity Properties}
\label{tab:inversedesign}
\begin{tabular}{lllrrr}
\toprule
 &  &  & Accuracy & Accuracy (all) & Error \\
Images? & Context & Model &  &  &  \\
\midrule
\multirow[t]{9}{*}{Yes} & \multirow[t]{3}{*}{5-NN} & NovaLite & 0.297 & 0.057 & 0.256 \\
 &  & NovaPro & 0.340 & 0.141 & 0.205 \\
 &  & NovaProFinetuned & 0.327 & 0.192 & 0.179 \\
\cline{2-6}
 & \multirow[t]{2}{*}{5-RN} & NovaLite & 0.400 & 0.062 & 0.213 \\
 &  & NovaPro & 0.364 & 0.115 & 0.208 \\
\cline{2-6}
 & \multirow[t]{4}{*}{0-Shot} & NovaLite & 0.500 & 0.010 & 0.168 \\
 &  & NovaLiteFinetuned & 0.389 & 0.305 & 0.140 \\
 &  & NovaPro & N/A & 0.000 & N/A \\
 &  & NovaProFinetuned & 0.370 & 0.311 & 0.179 \\
\cline{1-6} \cline{2-6}
\multirow[t]{8}{*}{No} & \multirow[t]{2}{*}{5-NN} & NovaLite & 0.278 & 0.114 & 0.277 \\
 &  & NovaPro & 0.308 & 0.156 & 0.195 \\
\cline{2-6}
 & \multirow[t]{2}{*}{5-RN} & NovaLite & 0.298 & 0.088 & 0.255 \\
 &  & NovaPro & 0.331 & 0.194 & 0.224 \\
\cline{2-6}
 & \multirow[t]{4}{*}{0-Shot} & NovaLite & 0.200 & 0.001 & 0.238 \\
 &  & NovaLiteFinetuned & 0.435 & 0.378 & 0.120 \\
 &  & NovaPro & 0.000 & 0.000 & 0.176 \\
 &  & NovaProFinetuned & 0.438 & 0.374 & 0.115 \\
\cline{1-6} \cline{2-6}
\bottomrule
\end{tabular}
\end{table}

\begin{table}
\caption{Elastic Property Prediction Accuracy}
\label{tab:inversedesign}
\begin{tabular}{lllrrr}
\toprule
 &  &  & Accuracy & MeanAccuracy (all)& Error \\
Inputs & Context & Model &  &  &  \\
\midrule
\multirow[t]{9}{*}{Code} & \multirow[t]{2}{*}{5-NN} & NovaLite & 0.21 & 0.21 & 0.29 \\
 &  & NovaPro & 0.24 & 0.24 & 0.28 \\
\cline{2-6}
 & \multirow[t]{3}{*}{5-RN} & NovaLite & 0.20 & 0.20 & 0.29 \\
 &  & NovaPro & 0.20 & 0.20 & 0.29 \\
 &  & NovaProFinetuned & 0.31 & 0.31 & 0.19 \\
\cline{2-6}
 & \multirow[t]{4}{*}{0-Shot} & NovaLite & 0.13 & 0.13 & 0.28 \\
 &  & NovaLiteFinetuned & 0.48 & 0.48 & 0.11 \\
 &  & NovaPro & 0.10 & 0.10 & 0.26 \\
 &  & NovaProFinetuned & 0.52 & 0.52 & 0.10 \\
\cline{1-6} \cline{2-6}
\multirow[t]{9}{*}{Images} & \multirow[t]{3}{*}{5-NN} & NovaLite & 0.21 & 0.21 & 0.29 \\
 &  & NovaPro & 0.24 & 0.24 & 0.25 \\
 &  & NovaProFinetuned & 0.40 & 0.40 & 0.13 \\
\cline{2-6}
 & \multirow[t]{2}{*}{5-RN} & NovaLite & 0.19 & 0.19 & 0.32 \\
 &  & NovaPro & 0.23 & 0.23 & 0.26 \\
\cline{2-6}
 & \multirow[t]{4}{*}{0-Shot} & NovaLite & 0.15 & 0.15 & 0.28 \\
 &  & NovaLiteFinetuned & 0.45 & 0.45 & 0.11 \\
 &  & NovaPro & 0.13 & 0.13 & 0.26 \\
 &  & NovaProFinetuned & 0.45 & 0.36 & 0.09 \\
\cline{1-6} \cline{2-6}
\multirow[t]{10}{*}{Images+Code} & \multirow[t]{3}{*}{5-NN} & NovaLite & 0.22 & 0.22 & 0.29 \\
 &  & NovaPro & 0.24 & 0.24 & 0.27 \\
 &  & NovaProFinetuned & 0.41 & 0.41 & 0.13 \\
\cline{2-6}
 & \multirow[t]{3}{*}{5-RN} & NovaLite & 0.18 & 0.18 & 0.33 \\
 &  & NovaPro & 0.19 & 0.19 & 0.28 \\
 &  & NovaProFinetuned & 0.40 & 0.40 & 0.12 \\
\cline{2-6}
 & \multirow[t]{4}{*}{0-Shot} & NovaLite & 0.13 & 0.13 & 0.27 \\
 &  & NovaLiteFinetuned & 0.44 & 0.44 & 0.11 \\
 &  & NovaPro & 0.14 & 0.14 & 0.26 \\
 &  & NovaProFinetuned & 0.54 & 0.54 & 0.08 \\
\cline{1-6} \cline{2-6}
\bottomrule
\end{tabular}
\end{table}

\begin{table*}[]
	\caption{Geometric and Material Property Similarity of Generated material reconstructions from images, text descriptions, and properties  (OLD DATA FROM ORIGINAL FINE TUNING EXPERIMENTS!!!)}
	\label{tab:generationresults}
	\scriptsize
	\begin{tabular}{llll|llllllllll}
\hline
\multicolumn{4}{c|}{\textbf{Method}} & \multicolumn{10}{c}{\textbf{Metric}} \\
\multicolumn{1}{c}{\multirow{2}{*}{\textbf{Model}}} & \multicolumn{1}{c}{\multirow{2}{*}{\textbf{Data Aug}}} & \multicolumn{1}{c}{\multirow{2}{*}{\textbf{Rep.}}} & \multicolumn{1}{c|}{\multirow{2}{*}{\textbf{Query}}} & \multicolumn{2}{c|}{\textbf{Geometric}} & \multicolumn{8}{c}{\textbf{Simulation}} \\
\multicolumn{1}{c}{} & \multicolumn{1}{c}{} & \multicolumn{1}{c}{} & \multicolumn{1}{c|}{} & IoU $\uparrow$ & \multicolumn{1}{l|}{Ch Dist.} & $A_UAI$ & $A_Z$ & $E$ & $G$ & $K$ & $\lambda$ & $\nu$ & $\rho$ \\ \hline
Nova-Lite & Seeds & Graph & 0-shot & 1.31e-2 & \multicolumn{1}{l|}{.0523} & 4.76e11 & 10.3 & 6.84e-4 & 2.78e-4 & 5.80e-4 & 4.04e-4 & \textbf{.311} & .0131 \\
Nova-Lite-FT & Seeds & Graph & 0-shot & \textbf{2.74e-2} & \multicolumn{1}{l|}{\textbf{.0225}} & 9.00e11 & 2.16 & 7.23e-4 & 2.94e-4 & 5.88e-4 & 4.06e-4 & .489 & \textbf{.0122} \\
GPT-4o & Seeds & Graph & 0-shot & 1.19e-2 & \multicolumn{1}{l|}{.0519} & \textbf{2.4e2} & \textbf{1.22} & \textbf{5.64e-4} & \textbf{2.38e-4} & \textbf{3.82e-4} & \textbf{2.23e-4} & .425 & .0154
\end{tabular}
\end{table*}

\subsection{Language Validation}

DSL can represent diverse materials.

DSL can be used for enumerative search.

\subsection{Material Understanding}

Property prediction from images

Property prediction from code

Property prediction from images + code

\subsection{Automated Design}

Reconstruction from images

Reconstruction from properties

Reconstruction from textual descriptions

Reconstruction from multi-modal combinations

\subsection{Baselines}

Are there any reasonable baselines to compare against, or is everything ablations?

\subsection{Ablations}

IR instead of DSL

Choice of VLM (Amzn, Llama, OpenAI)

Fine-Tuning, ICL: (random, nearest-neighbor), Zero-Shot, Fine-Tuning + ICL
}
% ====================================
%            Discussion
% ====================================
\section{Discussion, Limitations, and Future Work}
\label{sec:discussion}

Metamaterial design is an inherently multimodal, high-impact problem that requires complex reasoning and preference consideration, which makes it a natural test bed for AI development.
%interactive workflows, online or offline reinforcement learning (RL), and verified program synthesis. 
Conversely, metamaterial researchers have called for better data sets and AI-powered tools.
MetaDSL and MetaDB provide a common, traceable descriptor that both communities can adopt. 
As researchers contribute new designs in this format, the database will grow organically, giving machine-learning practitioners richer training data while delivering state-of-the-art design assistants to materials scientists.

Our current work provides a comprehensive framework toward these goals, offering myriad opportunities for improvement. We deliberately restricted our MetaAssist implementation to simple supervised fine tuning to provide a bedrock baseline for this new task. This provides common metric for techniques such as RAG to read papers and retrieve patterns, chain-of-thought reasoning to connect design intent to property profiles, and RL training with curriculum learning to generalize to novel inverse design profiles.% Remove this paragraph break if we need to claw back a few lines after editing

MetaDSL is designed to be retargetable (\Cref{supp:ecosystem}), but we currently only target a \procMeta backend that is more constrained than MetaDSL's design. A more flexible geometry kernel would unlock non-cubic and aperiodic tilings. Targeting a faster kernel would enable larger and more interactive workflows (e.g. interactive output simulation -- we currently often need multiple attempts to get a verifiably correct output), simulation-in-the-loop optimization, and an even-wider data-set scale. 

MetaDB also has ample opportunities for growth as a community project, including the implementation of additional generators \citep{Sun2023_PPL,Liu2022_PSL,AbuMualla2024_SymmetryInducedDesign,Makatura2023_ProcMeta}, systematic inclusion of singular design templates from metamaterial literature, and diversity-guided synthesis. Our program's explicit semantic structure could support taxonomy construction and intelligent exploration of large design spaces. With broad participation, MetaDB could become the primary resource for tracking metamaterial lineages, structure–property relationships, and mechanistic insights—paralleling the role ImageNet played in computer vision.

At the same time, our framework may be susceptible to misuse or misguided application -- particularly when it comes to our VLM-powered design assistant. Of course, our multilayer stack—simulation, code generation, and LLM reasoning—can introduce errors. This deserves particular attention in a domain like metamaterials, which is difficult to reason about intuitively, and an active frontier of science with rapidly changing understanding. The resulting materials may also be deployed in scenarios where inaccurate results may lead to catastrophic failure of engineered products or infrastructure. Thus, it is critical that each result must be validated before deployment, and communications should avoid overstating reliability.
Our format already takes small strides toward ensuring the accuracy and traceability of information by including detailed provenance records in each of our models. To further improve transparency, we also release our artifacts and the pipelines used to generate them. 
Moving forward, we believe it would be prudent to include additional safeguards such as automated validity checks, uncertainty estimates, safety factors, and optional gated access to high-fidelity simulators to reduce the risk of erroneous or unsafe designs.

\section{Conclusion}
\label{sec:conclusion}

We introduced \textbf{MetaGen}, a unified ecosystem for vision–language
metamaterial design that combines (i) \emph{MetaDSL}, a compact yet expressive
domain-specific language; (ii) \emph{MetaDB}, an over \numMaterialsInDatabaseRoundedDown-entry database with
paired geometry, renderings, and physics; (iii) \emph{MetaBench}, a
task-oriented benchmark that probes reconstruction, material understanding, and
inverse design; and (iv) \emph{MetaAssist}, the first VLM-driven CAD interface
for architected materials.  
Our baseline experiments illustrate that large
vision–language models offer promising performance for multi-modal translation and design generation.
Moreover, we provide a holistic vision for accelerated, symbiotic research at the intersection of
machine learning and architected materials.
With the introduction of MetaGen as both a challenging
benchmark for multimodal models and a practical toolkit for materials
scientists, our paper lays the foundation to bring this vision to life.

\begin{ack}
Beichen for providing training code for material gen (initial version)
Shiv, Rajiv, Brian, Nitin, Jack, Brittany, Viktor, <others?> at Amazon for funding, technical support and discussions

Amazon grant
MAD (overlapped with tail end of Liane's fellowship)

\end{ack}

%%%%%%%%%%%%%%%%%%%%%%%%%%%%%%%%%%%%%%%%%%%%%%%%%%%%%%%%%%%%
% References
% \section*{References} -- in the template, but not necessary when using bibliography instead of hardcoded
\bibliographystyle{unsrtnat}
\small{
    \bibliography{sources}
}

\newpage

\appendix

% ==================================
%     Section: MetaDSL
% ==================================
\section{Ecosystem Design}\label{supp:ecosystem}
The four components of the MetaGen ecosystem work together to achieve our design goals. We outline these goals and the design and organization decisions that achieve them here:

\begin{itemize}
    \item MetaDB
    \begin{itemize}
        \item Design Goals: Collect existing knowledge in a reconfigurable, reusable, and task independent manner
        \item Organization
        \begin{itemize}
            \item Primary Elements: Material Definitions; Provenance
            \item Derived Elements: Geometry; Computed Properties
        \end{itemize}
    \end{itemize}
    \item MetaBench
    \begin{itemize}
        \item Design Goals:
        \item Organization:
        \begin{itemize}
            \item Primary Elements: Structured Task Definitions; Target Data; References, Evaluation Procedures
            \item Derived Elements: Query Strings; Example Responses
        \end{itemize}
    \end{itemize}
    \item MetaDSL
    \begin{itemize}
        \item Design Goals: Eventual Comprehensiveness via Extensibility; Supports Hybrid Structures Easily; Ease of Use
        \item Design Decisions: Extensible Embedded Python DSL for extensibility and Ease-of-Us; Separation of Front-End Language from Geometry Kernel
    \end{itemize}
    \item MetaAssist
    \begin{itemize}
        \item Design Goals: Usable for general engineers; single interface across design silos; possibility of integrating unstructured data (literature, sketches, etc.)
        \item Elements: Interactive Interface; Trained Baseline Models
    \end{itemize}
\end{itemize}

\begin{figure}[h!]
    \centering
    \includegraphics[width=0.5\linewidth]{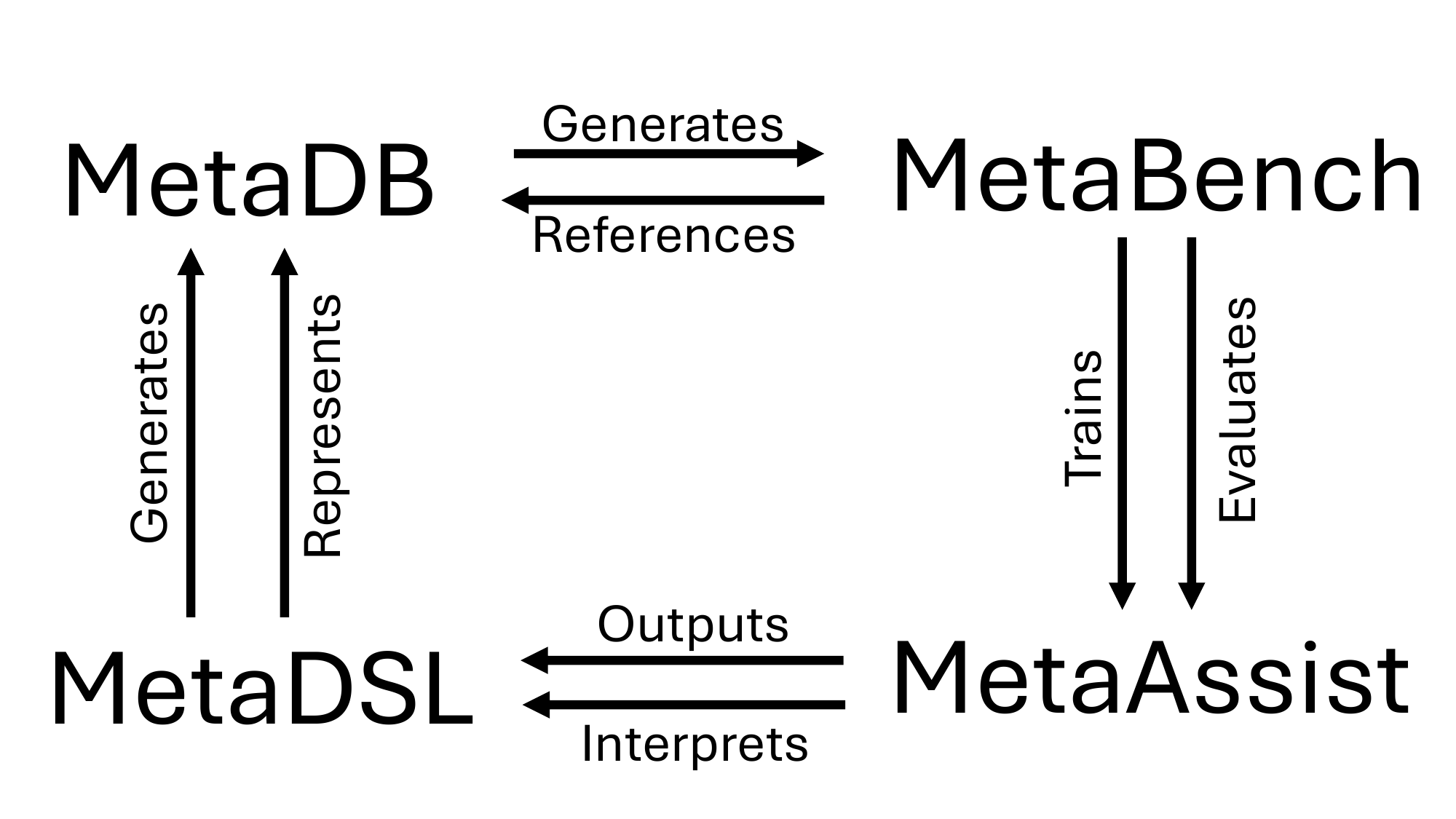}
    \caption{Relationships between MetaGen ecosystem components.}
    \label{fig:ecosystem}
\end{figure}

Each component supports the others, as illustrated in \Cref{fig:ecosystem}

\subsection{Ecosystem Development and Insights}
\label{sec:supp-eco-dev-and-insights}

The elements of this ecosystem were developed in concert with one another, going through 3 major iterations before arriving at their current state. MetaDSL was at the heart of each iteration, as the representation has a direct impact on the efficacy of the other three components:
\begin{itemize}
    \item MetaDB needs a representation that captures diverse structures, but also offers robust pathways for scalable (and, in this case, VLM-driven) structure generation, hybridization, mutation, sampling, etc.
    \item MetaBench can only be used for training and evaluation if it is built atop a large, diverse database.
    \item MetaAssist relies on a strong training corpus from MetaBench. MetaAssist also hinges on the intelligibility of the representation, and the model's ability to interpret, generate, and modify programs according to user input.
\end{itemize}

We defer the language-specific development details to \Cref{sec:supp-language-dev-insights}.

Outside the scope of the DSL, we also found that dataset management and curation posed a major hurdle. We improved diversity by continuously mining metamaterial literature for additional seed program designs. We expressed these seed programs as-parametrically-as-possible to allow for expert-driven sampling. As we scaled the dataset, we also realized that it would be critical to keep track of the programs’ sources and relationship to one another. This information is especially useful for navigation, contextualization and diversity management, particularly as the database grows in response to community effort. To manage this, we introduced a formalized provenance system for MetaDB.

%%%%%

% ====================================
%           MetaDSL
% ====================================

\section{MetaDSL}\label{supp:MetaDSL}

\subsection{Additional Implementation Details}\label{supp:MetaDSLImpl}

\begin{figure}
    \centering
    \includegraphics[width=0.8\linewidth]{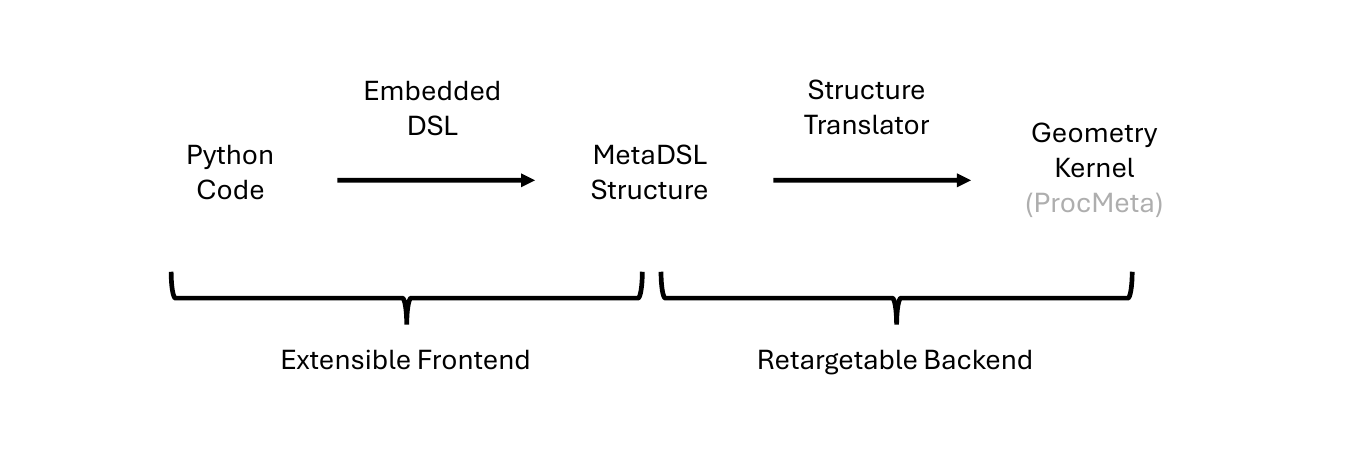}
    \caption{Overview of MetaDSL's implementation. MetaDSL programs are written in an embedded Python DSL frontend to allow for ease of use and extensibility. These structures are compiled into a structured intermediate representation, and a backend Translator converts these structures into geometry kernel instructions. In our implementation we used the geometry kernel from ProcMeta~\cite{Makatura2023_ProcMeta}. By separating the front-end representation from the backend geometry kernel, MetaDSL is flexible to both be extended in its frontend representation, and retargettable to different geometry backends for new applications, while keeping a compatible material representation.}
    \label{fig:metadslimpl}
\end{figure}

We implemented the core functionality of MetaDSL (version 1.1.0) with two goals in mind.
First, we wanted full support for the metamaterials that were expressible in our geometry kernel, \procMeta.
Second, we wanted our infrastructure to easily permit extensions in the future without invalidating existing programs.
We detail the current state of each feature category in our language: convex polytopes, skeletons, lifting procedures, tiles, and patterns.
For a full API description of the accessible functions, please refer to \Cref{sec:app-metadsl-api}. \Cref{fig:metadslimpl} shows an overview of the compiler architecture.

\paragraph{Convex Polytopes (CP)} Currently, all of our programs make use of three pre-defined CPs (as inspired by \procMeta): \dslFrag{cuboid}, \dslFrag{triPrism} and \dslFrag{tet}. 
The infrastructure to define custom convex polytopes exists, and most operators up to and including Tiles should generalize to such CPs. 
However, the patterning operations would need to be generalized before being able to operate on arbitrary CPs.

\paragraph{Skeletons}
Then, a \textit{skeleton} is constructed via a set of vertices and edges that are positioned relative to a common CP.
Each vertex is positioned on a particular CP entity (corner, edge, face, interior). 
Each CP entity is accessed via a semantically meaningful alias, permitting calls such as \eg \dslFrag{vertex(cuboid.edges.BACK\_LEFT)}. 
The \dslFrag{vertex} call also optionally takes a list \tInterp of interpolation values used to position the vertex within the entity.
If \tInterp is omitted, the returned point will be at the entity's midpoint (edge) or centroid (face/interior). 
Presently, corners ignore weights (since they cannot be moved); edges use linear interpolation; and faces use barycentric coordinates if they contain $3$ vertices or bilinear interpolation for quads.
If a CP with different polygonal faces (\eg pentagons) were implemented, an appropriate lower-dimensional vertex positioning specification would need to be devised.
Internally, the vertices are stored using weights over a full list of the CP corners, so additional specification interfaces can easily be defined. 

An ordered list of vertices can then be strung together into simple (non-branching, self-intersection-free) open or closed paths via the \dslFrag{Polyline} or \dslFrag{Curve} commands.
Each edge contained in a path infers and maintains information about its incidence on the CP -- including whether it is contained within a face, through the CP volume, coincident with a CP edge, \etc.
This is very useful when determining lifting function compatibility, as some procedures can only be applied when \eg every path edge is contained within a CP face.

Then, a \dslFrag{skeleton} is used to combine a set of vertices or polylines/curves into a larger, more complex element, over which additional organizational information is computed. 
Skeletons infer the connected components formed by the inputs, then categorize them based on their topology.
Thus, a skeleton may be labeled as a simple closed loop, even if the input is a set of open paths.
Again, these insights are critical for determining the skeleton's compatibility with downstream operations, such as lifting procedures.
We also included infrastructure for the skeletons to infer and track their total incidence on each entity of the reference CP, including the dimensionality (\eg point or line) of an intersection -- however, this feature is not fully implemented in the current MetaDSL version.

\paragraph{Lifting Procedures}
\textit{Lifting procedures} are used to transform the skeleton into a volumetric object.
Simple procedures like \dslFrag{Spheres} instantiate a sphere of the given radius centered at each vertex in the skeleton.
Similarly, \dslFrag{UniformBeams} instantiates a beam of the given thickness centered along each path of the input skeleton.
The shell operators (\dslFrag{UniformDirectShell}, \dslFrag{UniformTPMSShellViaMixedMinimal}, and \dslFrag{UniformTPMSShellViaConjugation})  solve for a surface that spans the provided boundary curve before expanding the surface to the desired thickness. 
Our shell and beam procedures mimic those defined by \procMeta, as they cover a wide range of metamaterial classes and were already (by construction) natively supported by our geometry kernel. 
Our \dslFrag{Curve} and \dslFrag{Polyline} commands correspond to their smooth/non-smooth edge chains, respectively.
Unlike the original, we chose to explicitly separate several operators that were previously lumped together, which clarified and minimized the number of exposed parameters for each call.

\paragraph{Tiles}
To create an embedded, patternable tile, we provide a list of one or more lifted skeletons as input to the \dslFrag{Tile} operator. The tile operator also takes as input the embedding information, which will be used to embed the CP and, in turn, each vertex of the contained skeleton(s). To obtain the embedding information, each CP implements at least one \dslFrag{embed} function, which takes high level parameters such as the min/max position of the CP's AABB.

Because of constraints imposed by $\procMeta$ -- that these must form a partition of the unit cell -- our code currently treats these CPs with some additional assumptions. 
Specifically, though the cuboid need not be a cube, it must have right angles everywhere, and edge lengths must be $1/2^k$ for some positive integer $k$; in practice, $k \in [1,..4]$. 
The triPrism is assumed to be an isoceles triangle with a right angle.
The tet similarly has a base that is an isoceles triangle with a right angle, and a fourth vertex that is located directly above one of the $45$ degree angles.
These assumptions would ideally be relaxed in a future version of MetaDSL.

\paragraph{Patterns}
Patterns are currently the most restricted feature of MetaDSL, as we restrict our dataset to programs that can be compiled down to the language and solver set described by \procMeta. 
Thus, rather than extending our structures to a more arbitrary tiling in $\R^3$, all of our structures have a translational unit residing in a unit cube.
The pattern operators were written in a way that allows for additional, extended tiling procedures.
We prioritized mirrors, because they are sufficient to express a wide range of common metamaterial designs, and they are often used in generative metamaterial design schemes, as the connectivity requirements are simpler than most other operations. 
We also have limited support for other operations such as \dslFrag{Rotate180} and \dslFrag{Translate}, which can be used inside the \dslFrag{Custom} pattern specifier.
Currently, these limited operations are only defined for specific transformations on cuboids.
We look forward to an expanded MetaDSL that includes full support for these patterning operations, at least over the pre-built CPs that currently exist.
In the long term, we envision a patterning system that extends well beyond this, to support large, potentially aperiodic or asymmetric tilings composed of one or more tiles with arbitrary CPs. 
This is a very difficult problem, and will itself present an interesting set of research directions, including how to intuitively specify these patterns and how to characterize their compatibility/validity.

\subsection{Example Programs}\label{supp:ExamplePrograms}
\begin{figure*}
    \centering
    \begin{minipage}{0.7\textwidth}
    \centering
    \begin{lstlisting}[language=Python]
from metagen import *

def make_structure(shell_thickness=0.03) -> Structure:
    v0 = vertex(tet.edges.BOTTOM_LEFT)
    v1 = vertex(tet.edges.TOP_LEFT)
    v2 = vertex(tet.edges.TOP_RIGHT)
    v3 = vertex(tet.edges.BOTTOM_RIGHT)

    c0 = Curve([v0, v1, v2, v3, v0])

    skel = skeleton([c0])
    shell = UniformTPMSShellViaConjugation(skel, shell_thickness)

    embedding = tet.embed(0.5)
    tile = Tile([shell], embedding)
    pat = TetFullMirror()
    obj = Structure(tile, pat)

    return obj
    \end{lstlisting}
    \end{minipage}
    \begin{minipage}{0.29\textwidth}
    \centering
    \includegraphics[width=\linewidth]{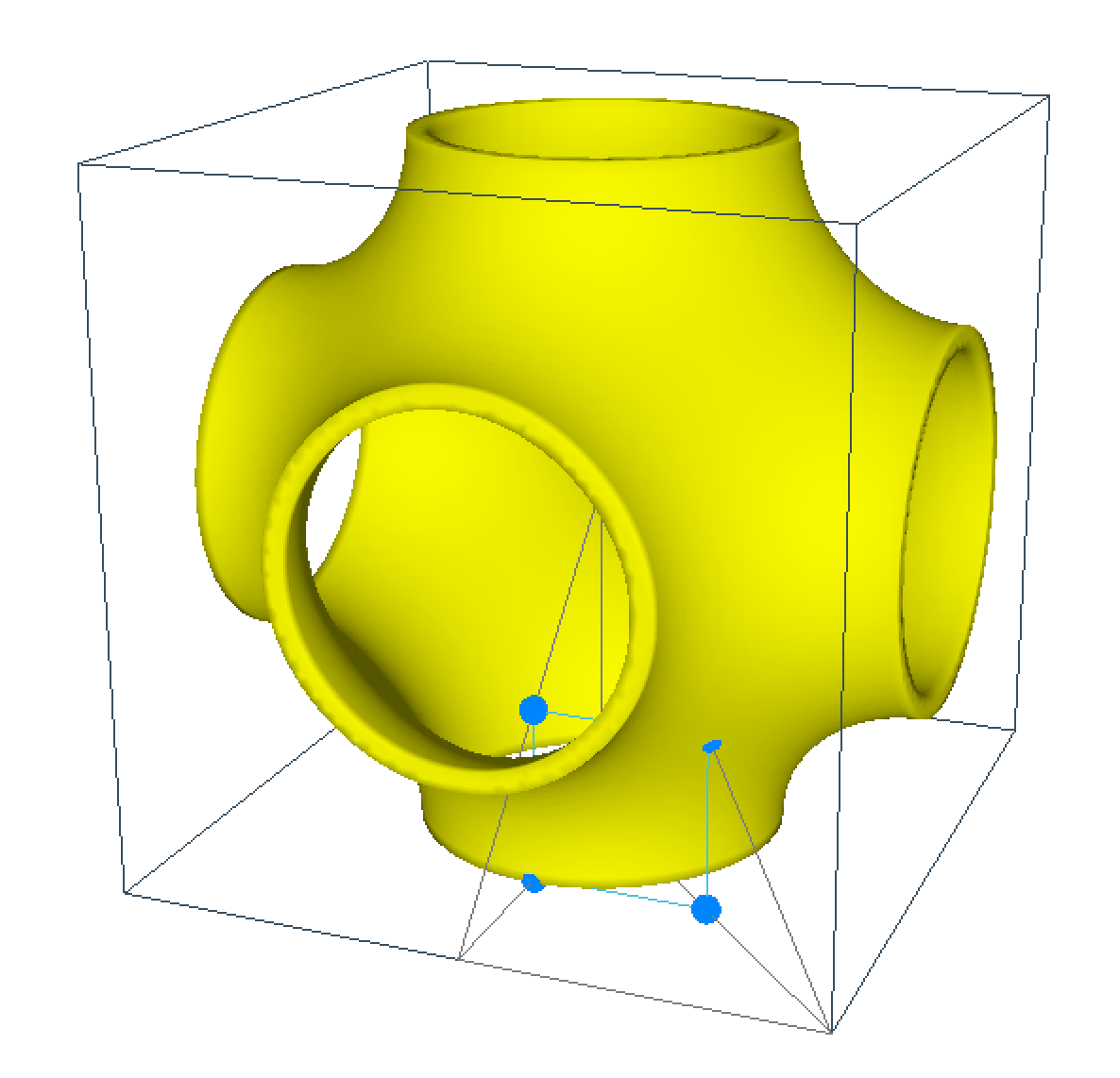}
    \end{minipage}
    \caption{Example program and corresponding geometry for the Schwarz P structure. }
    \label{fig:app-example-program-schwarzp}
\end{figure*}

\begin{figure*}
    \centering
    \begin{minipage}{0.8\textwidth}
    \centering
    \begin{lstlisting}[language=Python]
from metagen import *

def make_structure(beamRadius_narrow=0.03, beamRadius_wide=0.1) -> Structure:
    embed = cuboid.embed(0.5, 0.5, 0.5, 
                    cornerAtAABBMin=cuboid.corners.FRONT_BOTTOM_LEFT)
    
    v0 = vertex(cuboid.corners.FRONT_BOTTOM_LEFT)
    v1 = vertex(cuboid.corners.BACK_TOP_RIGHT)
    p0 = Polyline([v0, v1])

    skel = skeleton([p0])
    liftedSkel = SpatiallyVaryingBeams(skel, [[0, beamRadius_narrow], 
                                              [0.5, beamRadius_wide],
                                              [1, beamRadius_narrow]])

    tile = Tile([liftedSkel], embed)
    pat = Custom(Rotate180([cuboid.edges.BACK_RIGHT, 
                            cuboid.edges.BACK_LEFT], True,
                        Rotate180([cuboid.edges.TOP_RIGHT], True)))
    obj = Structure(tile, pat)

    return obj
    \end{lstlisting}
    \end{minipage}
    \begin{minipage}{0.19\textwidth}
    \centering
    \includegraphics[width=\linewidth]{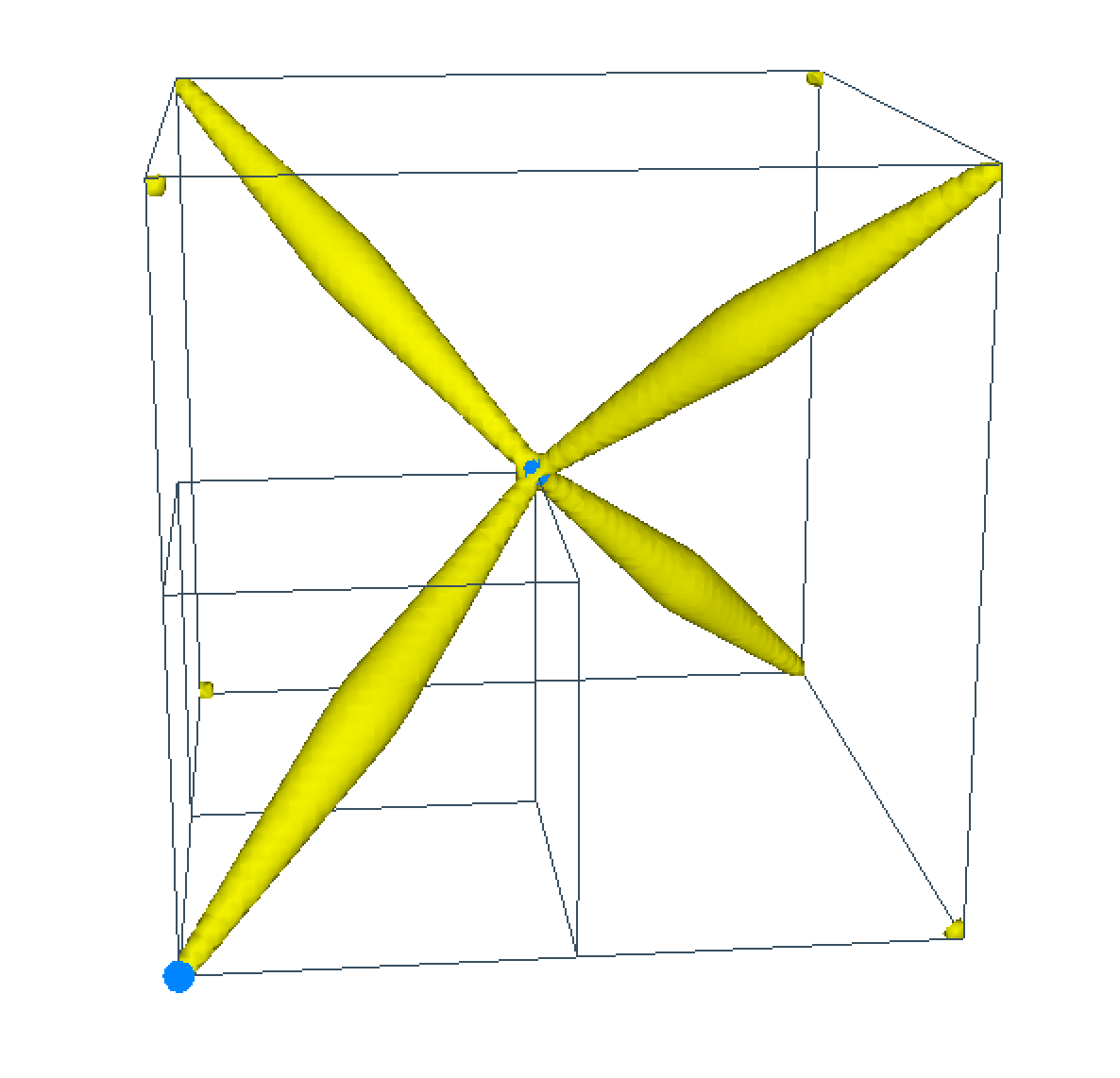}
    \end{minipage}
    \caption{Example program and corresponding geometry for the pentamode structure. }
    \label{fig:app-example-program-pentamode}
\end{figure*}

Example program-structure pairs are listed in \Cref{fig:app-example-program-schwarzp} and \Cref{fig:app-example-program-pentamode}. Many additional models can be found in the accompanying data. 

\subsection{MetaDSL vs. ProcMeta}
\label{sec:app-metadsl-vs-procmeta}
As suggested by \Cref{sec:supp-eco-dev-and-insights} and the architecture diagram in \Cref{fig:metadslimpl}, MetaDSL is distinct from and strictly more general than ProcMeta, with a design philosophy all its own.
Our approach was motivated by our early experiments with ProcMeta, which revealed a critical shortcoming: important information was represented implicitly in the ProcMeta GUI interface, and was entirely absent from the ProcMeta graph representation. To make this information accessible to LLMs (and more easily accessible to humans), we implemented a programmatic interface, MetaDSL, that compiles to the same geometry kernel as ProcMeta, but provides several practical advantages (see \Cref{tab:supp_metaDSL_vs_procmeta}).

Most importantly, MetaDSL introduces explicit, referenceable bounding volumes (BVs), which are critical for verifying and enforcing the preconditions of geometry operations. In the ProcMeta GUI, BVs exist only as non-referenceable visual aids; users must manually align coordinates, and no automated compatibility checks are possible. ProcMeta graphs omit BVs entirely. MetaDSL represents BVs through a CP abstraction, which enforces constraints by construction, enables type checking, and cleanly separates tile content from patterning, improving modularity and reconfigurability. These features align the representation more closely with the valid shape space, aiding both human designers and LLMs in producing valid, diverse structures.
MetaDSL programs also make heavy use of programmatic features absent from ProcMeta graphs. Semantic variable names, comments (avg. 4/program), and parametric variables improve human interpretability and support natural-language reasoning for LLMs. Loops and helper functions are also common, appearing in 1,744 and 2,103 of the 13,284 core programs respectively. These features allow compact, self-consistent definitions that would be unwieldy if unrolled or inlined into a ProcMeta graph.

We tested LLM-based augmentation using ProcMeta JSON instead of MetaDSL. MetaDSL yielded: (1) higher code validity (75\% vs. 54\%), (2) more structurally focused reasoning rather than boilerplate handling, and (3) lower token usage (580 vs. 1,049 tokens on average for o4).
Beyond these immediate benefits for LLM usage and dataset generation, our DSL interface also makes MetaDSL a more flexible platform from which to build further extensions, which facilitates its intended purpose as the seed of a wider community project.

\paragraph{Extensibility}
The MetaDSL interface naturally generalizes to shape spaces that would be difficult to represent in ProcMeta’s graph approach. For example, implicit functions are common in metamaterial design, but they would be cumbersome to represent in ProcMeta’s graph. However, MetaDSL could naturally include them: rather than an explicit Skeleton, we could use the implicit function to define a SkeletonGenerator; this could then be fed to an Implicit lifting function, which would solidify a given isovalue range. Non-trivial patterning would also be possible through MetaDSL’s Custom pattern interface. For example, given a set of mutually compatible unit cells (like the left/right faces of \Cref{fig:seed_variants}a,b,c,f), simple translations could combine them into an elongated, interleaved tile (e.g. ABCCBA). With enhanced compatibility determination, we could also create Pattern procedures for scholastic or aperiodic tilings. This will allow MetaDSL to expand alongside developments in metamaterial design.

\begin{table}[]
    \centering
    \scriptsize
    \renewcommand{\arraystretch}{1.4}% for the vertical padding
    \begin{tabular}{p{2.5cm}p{5.5cm}p{5.5cm}}
        \hline
         &MetaDSL  &ProcMeta\\
        \hline
         Compactness
         & \textbf{Shorter}, less boilerplate. Easier to read, less likely to exceed token limits  
         & Longer, more boilerplate. Exceeds context of small, lightweight models.\\
         Modules
         & \textbf{Highly reusable}. Patterns defined in composable chunks (eg TetMirror), independent of tile contents. Skeletons defined independent of embedding, easily scale to different Tiles.
         & \textbf{No support.} Limited reuse. Patterns can’t exist independently; no pre-built Patterns. Absolute Skeletons, cannot easily be rescaled. \\
         Relative vs. Absolute Positioning
         & Positions and transforms use \textbf{local coordinates} (\ie [0,1]) wrt named entities (\verb|cuboid.edges.TOP_LEFT|) in abstract polytopes. Robust for generation, clear design space bounds, more intuitive.
         &Positions and transforms use \textbf{absolute coordinates}. Easily misaligned, difficult to visualize without plotting. Unsuitable for VLMs, which struggle with computation/spatial tasks. \\
         BV representation
         &\textbf{Explicit BV with named, referenceable entities.} Facilitates verifiable parametric design, \eg, vertex constrained to given BV edge. Allows type/error checking.
         &\textbf{Implicit or Absent BV}: drawn as a visual aid in the GUI, but not represented/preserved in the graph. Never referenceable. \\
         Type/Error checking
         &\textbf{Type/incidence tracking to ensure compatibility} -- \eg conjugate TPMS require a closed loop where every edge lies in a BV face, and every BV face contains at least 1 loop edge. This is known from our representation and verified by downstream operations. Helps determine valid substitutions for mutations, even when large changes are proposed, leading to greater diversity. Critical for complex patterning, to determine compatibility of proposed-adjacent faces.
         &\textbf{None.} The burden of verification (for \eg vertices on BV edges or edges in BV faces) is left to the user -- infeasible for agentic design. Bad inputs crash ProcMeta with no explanation or suggested improvements.\\
         Simplified Operations
         &\textbf{Abstractions simplify element creation}; \eg, Sphere() takes a center point and a radius, as one would expect. Easier for humans and LLMs.
         &Strict compliance with the given graph interface makes \textbf{some operations cumbersome}; \eg for a sphere, thicken a 0-length edge chain over 2 co-located vertices\\
         Semantic information
         &\textbf{Complete support.} Comments and meaningful variable names improve readability and admit metadata (provenance, parameter bounds)
         &No support. \\
         Parameters
         &\textbf{Complete support.} Allows parametrized models and family generators.
         &\textbf{None.} Explicit positions etc. only. Variations defined as separate graphs. Difficult/impossible to infer constraints or design space from the graph description.\\
         Loops, Functions
         &\textbf{Supports complex logic} that would be tedious to implement otherwise. Functions are especially useful for hybridization, as programs can be directly reused and/or rescaled.
         &\textbf{No support.} Each instance must be created/connected individually. Even hybridization is difficult, because subgraphs cannot be inserted directly -- the identifier/references of each node must be updated.\\
        \hline
    \end{tabular}
    \caption{Detailed differences between the interfaces for MetaDSL and ProcMeta.}
    \label{tab:supp_metaDSL_vs_procmeta}
\end{table}

\subsection{Language Development Process and Insights}
\label{sec:supp-language-dev-insights}

As mentioned in \Cref{sec:supp-eco-dev-and-insights}, our geometry representation went through 3 major stages.

In the first iteration, we represented metamaterials using ProcMeta graphs directly. This had several issues: it was not compact enough for the context windows of small, lightweight models; intuitiveness and editability suffered dramatically without the aid of a GUI editing tool; the graphs’ use of absolute coordinates proved challenging for LLMs (which struggle with spatial reasoning); and the program manipulations (e.g. hybridization, mutation) were unwieldy and fragile, with low validity rates that prohibited effective dataset scaling and diversification. This limited the breadth of MetaDB and MetaBench, while curtailing the efficacy of MetaAssist. 

To address this, we designed a higher-level language that became MetaDSL-v0. This approach had a compact, modular, bilevel design that was embedded within Python and thus permitted semantically meaningful content; as such, it solved the context length and human editability issues of ProcMeta. It allowed for relative positioning, which mitigated the issues with coordinates while improving components’ reusability. It also allowed for dataset augmentation through programmatic mutation, and improved the efficacy of VLM-based hybridization and mutation -- we attributed this jump to our Python embedding, as VLMs show great facility with Python. Still, MetaDSL-v0 remained fragile: generated programs frequently failed, and database augmentations showed limited diversity.

Analysis of MetaDSL-v0’s failure modes offered several insights; we arrived at the current MetaDSL by addressing each in turn. First, we noticed that VLMs often used hallucinated synonyms, such as \verb|TOP_LEFT| vs \verb|LEFT_TOP|; we added overloads for all reasonable variations of our functions and attributes. We also found that it was critical to abrogate as much spatial reasoning from the VLM as possible: a full 1/3 of failures were due to the VLM’s improper positioning of vertices that form the concrete polytope tiles. We circumvented this through abstracted tile embedding functions, which generate valid embeddings from simple, meaningful parameterizations. In our final large-scale change, we swapped the relative order of lifting functions and tile embeddings (previously Embed then Lift; now, Lift then Embed). This change improved the modularity and compositionality while reducing verbosity -- for example, this change allows multiple skeletons to reside in a shared Tile embedding, such that they can be patterned as a single unit. This change also paved the way for patterning of more diverse geometry-generation methods in future extensions. As a result, MetaDSL showed dramatic improvements in generation/mutation rates, and -- in turn -- significantly more diverse LLM-driven hybridizations.

% ==================================
%     Section: MetaDB
% ==================================
\section{MetaDB}\label{supp:MetaDB}

\subsection{Database Layout}\label{supp:MetaDBLayout}

MetaDB is structured into 4 primary directories:
\begin{itemize}
    \item literature: Literature references that are the sources for hand-authored models.
    \item models: MetaDSL programs and their outputs.
    \item generators: Programs that create and augment models
    % \item metadata: Dataset statistics and metadata
    \item benchmark: The MetaBench benchmark
\end{itemize}

Data items in MetaDB can reference other items by path. These paths are either absolute (start with a forward slash ``/'') or relative (no leading slash). Absolute paths are assumed to start at the root of the database structure. For example, a model may reference the paper that defined it in its sources as \verb|/literature/...|.

\subsection{Provenance Information}\label{supp:MetaDBProvenance}

Each Model in MetaDB starts with a triple-single-quote (\verb|'''|) delimited yaml string called the header-block. This contains useful metadata about the program, including provenance information about how it was created, and what sources it draws on. Provenance information is recorded in two places in the header block.

The primary location is in the ``sources'' key. This is a dictionary where the keys are MetaDB paths to literature, models, or generators that are the source of this model. The secondary location is in \verb|file_info|$\to$\verb|generator_info|. For models that are autogenerated via enumeration or augmentation this section contains a MetaDB path to the script that generated the file, the arguments that were passed into that script, and specific \verb|structure_details| that specified this particular model.

\subsection{Hybridization Implementation}\label{supp:MetaDBHybridization}
We hybridized hand-authored models using calls to OpenAI's o4-mini model using a reasoning effort of "medium". For every pair and triplet of authored models, we used the following prompt template:

\begin{lstlisting}[breaklines=True]
You have access to a DSL whose specification is as follows:
{api_description}

I want you to help discover unique new programs. Do this by genetic crossover based on these parent Metagen DSL programs:

1)
```python
{program 1 code}
```

2)
```python
{program 2 code}


Combine relevant structural/logical features from each sample into one coherent DSL program.
Be sure to:
- Respect the DSL syntax strictly. 
- Maintain correctness in the final structure definition.
- Keep the final program well-formed and ready to be run as a standard Metagen DSL generator.
- Provide minimal descriptive comments.

Return only the resulting code in a single code block.
\end{lstlisting}

where \verb|api_description| is the MetaDSL API specification given in \Cref{sec:querytemplates}, and the program code is listed excluding the header block.

\subsection{Mutation Implementation}\label{supp:MetaDBMutation}
Our mutation script loads a DSL model from file and constructs the corresponding Structure object in memory. Then, it is able to modify the structure along 4 different axes.
Two of the axes allow discrete adjustments:
(1) switching any \dslFrag{Polyline} to a \dslFrag{Curve} or vice versa; and
(2) selecting a different lifting procedure from the set of options compatible with the skeleton (as inferred by our type system).
The remaining modification axes permit continuous variations: 
(3) repositioning a vertex within its CP element;
and
(4) selecting a different thickness specification for any lifting procedures.
To generate a given variant, each modification axis was permitted with a pre-specified probability; we used $\Pr=0.7$ for both discrete changes, $\Pr=0.9$ for vertex perturbation, and $\Pr=0.98$ for thickness perturbation.
Once a given perturbation category was permitted, we looped over each opportunity for said modification within our structure specification, and evaluated a random number against the same respective probability to decide whether this specific instance should be modified or not.
For example, with $\Pr=0.7$ we allow \dslFrag{Polyline}/\dslFrag{Curve} swaps in the variant; then, each time a candidate \dslFrag{Polyline}/\dslFrag{Curve} is identified, we enact the swap with $\Pr=0.7$.
Once an instance has been approved, the specific replacement value was chosen at random from the appropriate set of options (if more than one available).
The updated structure is then written to file using the \dslFrag{dslTranslator}, which writes a DSL model from a Structure object.
Additional mutation procedures could be implemented to further increase the vawriety of resulting structures.

Provenance Information is stored in the \verb|sources| section of each program's header block. This is a dictionary where the keys are database paths.

\subsection{Material Properties}
\label{sec:app-database-properties}

Our simulation provides the $6\times 6$ elastic tensor $C$ in Voigt notation, along with the compliance matrix, $S = C^{-1}$. From this, we extract $18$ common material properties:
\begin{itemize}
    \item $E$: Young's Modulus, Voigt-Reuss-Hill (VRH) average, relative to $E_{\textrm{base}}$.  
    \item $E_1,E_2,E_3$: Directional Young's Moduli, relative to $E_{\textrm{base}}$
    \item $G$: Shear Modulus (VRH average), relative to $E_{\textrm{base}}$
    \item $G_{23},G_{13},G_{12}$: Directional Shear Moduli, relative to $E_{\textrm{base}}$
    \item $\nu$: Poisson ratio (VRH average)
    \item $\nu_{12}, \nu_{13}, \nu_{23}, \nu_{21}, \nu_{31}, \nu_{32}$: Directional Poisson ratios
    \item $K$: Bulk modulus (VRH average), relative to $E_{\textrm{base}}$
    \item $A$: Anisotropy (universal anisotropy index)
    \item $V$: Volume Fraction. 
\end{itemize}

\subsection{Ensuring MetaDB Quality}
\label{supp:metadb-quality}

MetaDB is founded on a strong basis of expert programs, including 50 hand-authored examples sourced from diverse, singularly-developed designs in metamaterial literature.
This large, diverse collection of seeds is unique to MetaDB, as most large datasets are derived exclusively from a small set of procedural generators. For example, \citet{xue2025_mindmicrostructureinversedesign} creates a database of 180k samples, 78\% of which stem from variations of the topologies in Elastic Textures \citep{Panetta2015_ElasticTextures}. The remaining 22\% stem from similar generators for planar- and curved-shell structures \citep{Liu2022_PSL, Sun2023_PPL}. Because of the reliance on such generators, \citet{xue2025_mindmicrostructureinversedesign} does not offer any representation of \eg CSG-style structures like the Bucklicrystal of \citet{babaee20133d}. However, the bucklicrystal is part of our database, as shown in \Cref{fig:metadb-assortment}(i), center). MetaDB also already includes Elastic Textures, and similar generators could be implemented for the remaining sources mentioned above.

To ensure that MetaDB only contains high-quality material definitions -- even when automatically generating a large portion of our entries -- material models are only added after they have passed a series of basic checks. Presently, this includes 3 criteria:
\begin{itemize}
    \item \textbf{MetaDSL compilation:} the model must contain valid python code that successfully evaluates to a MetaDSL Structure object. This includes all runtime type checking done by MetaDSL. 
    \item \textbf{Valid Geometry Generation:} after the MetaDSL Structure object is transpiled into the target geometry kernel (in our case, ProcMeta), the kernel is run. We check the resulting geometry for validity, as measured by a non-null result that is tilable in 3D. To determine tilability, we tile the base cell in a $3\times3\times3$ lattice, then check that the boundaries are periodic and that at least one connected component of this larger base cell reaches all boundaries.
    \item \textbf{Physically Consistent Simulation Results:} the simulator must return reasonable results that obey physical constraints. For example, since our simulation is normalized by the base material's Young's modulus $E_{\textrm{base}}$, it must be the case that our simulation returns $E \leq 1$.
\end{itemize}

\iffalse % started writing this then found it already exists
\subsection{Task Format}\label{supp:taskformat}
MetaBench tasks are stored as JSON records with the following fields:
\begin{itemize}
    \item task\_category: One of reconstruction, inverse\_design, or material\_understanding
    \item task\_type: Fine-grained category (e.g. 1\_view\_reconstruction)
    \item label: A unique string identifier
    \item source: MetaDB reference of the model this task is derived from. This is optional, as there could be no source for, e.g. inverse design tasks, but all tasks in the initial MetaBench release do have sources
    \item data: A dictionary of target data for evaluation. This is task category/type dependent. Target data can be stored directly in the JSON (used for compact data like target properties), or as MetaDB references (used for larger and non-textual data like images and voxel grids).
    \item query: An example string query 
\end{itemize}
\fi

% ==================================
%     Section: Results
% ==================================
\section{Further Benchmark Results}\label{supp:extrabench}

\subsection{Expanded Quantitative Results}
In this section we extend the primary table from the paper \Cref{tab:benchmarksummary} to include 95\% confidence intervals, computed using the standard-error approximation (\Cref{tab:confidenceintervals}). We also show more detailed tables for each task category, broken down to the individual task type. These extended views do not change the primary observations from the main text, but do highlight the differences between subtasks.
\begin{table}[]
    \centering
    \scriptsize
    \begin{tabular}{cccccccc}
    \toprule
    Category & \multicolumn{2}{c}{Inverse Design} & \multicolumn{2}{c}{Material Understanding} & \multicolumn{3}{c}{Reconstruction} \\
    Metric & Error & Valid & Error & Valid & CD & IoU & Valid \\
    Model &  &  &  &  &  &  &  \\
    \midrule
    LLaVAOmniTask & \textbf{0.011 ± 0.002} & \textbf{91.9\% ± 0.9\%} & 0.024 ± 0.004 & \textbf{100.0\% ± 0.0\%} & 0.034 ± 0.001 & 0.490 ± 0.008 & 82.9\% ± 0.9\% \\
    LLaVASingleTask & 0.036 ± 0.007 & 81.9\% ± 3.2\% & \textbf{0.018 ± 0.004} & \textbf{100.0\% ± 0.0\%} & \textbf{0.029 ± 0.003} & \textbf{0.524 ± 0.030} & 83.8\% ± 3.2\% \\
    NovaLite & 0.060 ± 0.023 & 2.7\% ± 0.6\% & 0.200 ± 0.005 & \textbf{100.0\% ± 0.0\%} & 0.119 ± 0.003 & 0.051 ± 0.003 & 19.3\% ± 0.9\% \\
    NovaOmniTask & 0.026 ± 0.002 & 91.4\% ± 1.0\% & 0.032 ± 0.005 & \textbf{100.0\% ± 0.0\%} & 0.045 ± 0.001 & 0.334 ± 0.007 & \textbf{87.2\% ± 0.8\%} \\
    NovaSingleTask & 0.032 ± 0.007 & 79.2\% ± 3.4\% & 0.153 ± 0.006 & \textbf{100.0\% ± 0.0\%} & 0.059 ± 0.003 & 0.205 ± 0.020 & 84.8\% ± 3.2\% \\
    OpenAIO3 & 0.038 ± 0.006 & 24.7\% ± 1.5\% & 0.077 ± 0.005 & \textbf{100.0\% ± 0.0\%} & 0.053 ± 0.001 & 0.147 ± 0.004 & 54.6\% ± 1.1\% \\
    \bottomrule
    \end{tabular}
    \caption{Benchmark summary with confidence intervals.}
    \label{tab:confidenceintervals}
\end{table}

\paragraph{Significance} In \Cref{tab:confidenceintervals} we show 95\% confidence intervals around the sample means for our top-level task categories. From these we can see that for every task that LLaVASingleTask outperformed LLaVAOmniTask, the confidence intervals actually overlap, indicating that this performance boost from single-task training may not be significant. This reaffirms our decision to base our metamaterial co-pilot on the OmniTask trained models.

\begin{table}[]
    \centering
    \scriptsize
    \begin{tabular}{ccccccccccccc}
    \toprule
        Task & \multicolumn{3}{c}{1 View} & \multicolumn{3}{c}{2 View} & \multicolumn{3}{c}{3 View} & \multicolumn{3}{c}{4 View} \\
        Metric & CD & IoU & Valid & CD & IoU & Valid & CD & IoU & Valid & CD & IoU & Valid \\
        Model &  &  &  &  &  &  &  &  &  &  &  &  \\
        \midrule
        LLaVAOmniTask & \textbf{0.036} & \textbf{0.458} & 82.3\% & \textbf{0.033} & \textbf{0.497} & 83.0\% & \textbf{0.032} & \textbf{0.509} & 83.2\% & 0.033 & 0.497 & 83.2\% \\
        LLaVASingleTask & \textemdash{} & \textemdash{} & \textemdash{} & \textemdash{} & \textemdash{} & \textemdash{} & \textemdash{} & \textemdash{} & \textemdash{} & \textbf{0.029} & \textbf{0.524} & 83.8\% \\
        NovaLite & 0.119 & 0.049 & 18.7\% & 0.117 & 0.050 & 17.0\% & 0.118 & 0.053 & 22.0\% & 0.125 & 0.050 & 25.0\% \\
        NovaOmniTask & 0.047 & 0.307 & \textbf{87.5\%} & 0.044 & 0.338 & \textbf{87.5\%} & 0.043 & 0.350 & \textbf{86.2\%} & 0.044 & 0.346 & \textbf{87.8\%} \\
        NovaSingleTask & \textemdash{} & \textemdash{} & \textemdash{} & \textemdash{} & \textemdash{} & \textemdash{} & \textemdash{} & \textemdash{} & \textemdash{} & 0.059 & 0.205 & 84.8\% \\
        OpenAIO3 & 0.052 & 0.150 & 36.8\% & 0.055 & 0.141 & 58.9\% & 0.052 & 0.151 & 62.6\% & 0.052 & 0.155 & 68.5\% \\
        \bottomrule
    \end{tabular} 
    \caption{Reconstruction Results Broken Down by task type.}
    \label{tab:reconstruction}
\end{table}

\paragraph{Categorical Results} Tables \ref{tab:reconstruction}, \ref{tab:invdesign}, and \ref{tab:understanding} break down \Cref{tab:benchmarksummary} for each task category into its task variations (number of views, targets, etc.). These provide a more even point of comparison between single and omni-task models because the results are aggregated over exactly the same examples. By contrast, in the primary table, the omni-task models are averaging over more and different tasks; thus, they may be biased by overall easier or harder requests.

In reconstruction (\Cref{tab:reconstruction}), we see a trend that having more viewpoints makes reconstruction slightly easier. We can see that the inclusion of these harder tasks did pull down the OmniTask average slightly in the general benchmark, but it was not the deciding factor. A similar trend is seen in Nova, but there the gap is significantly larger.

\begin{table}[]
    \centering
    \tiny
    \begin{tabular}{ccccccccccccc}
    \toprule
    Task & \multicolumn{2}{c}{1 Target} & \multicolumn{2}{c}{2 Target} & \multicolumn{2}{c}{3 Target} & \multicolumn{2}{c}{4 Target} & \multicolumn{2}{c}{5 Target} & \multicolumn{2}{c}{6 Target} \\
    Metric & Error & Valid & Error & Valid & Error & Valid & Error & Valid & Error & Valid & Error & Valid \\
    Model &  &  &  &  &  &  &  &  &  &  &  &  \\
    \midrule
    LLaVAOmniTask & 0.023 & \textbf{99.0\%} & \textbf{0.011} & \textbf{94.3\%} & \textbf{0.007} & \textbf{93.1\%} & \textbf{0.010} & 89.7\% & \textbf{0.008} & 88.3\% & \textbf{0.008} & 87.9\% \\
    LLaVASingleTask & \textemdash{} & \textemdash{} & \textemdash{} & \textemdash{} & \textemdash{} & \textemdash{} & 0.036 & 81.9\% & \textemdash{} & \textemdash{} & \textemdash{} & \textemdash{} \\
    NovaLite & 0.036 & 2.1\% & 0.049 & 4.6\% & 0.043 & 2.0\% & 0.078 & 3.2\% & 0.083 & 1.2\% & 0.072 & 2.8\% \\
    NovaOmniTask & \textbf{0.020} & 90.3\% & 0.018 & 90.6\% & 0.024 & 90.5\% & 0.029 & \textbf{92.7\%} & 0.035 & \textbf{90.2\%} & 0.028 & \textbf{94.0\%} \\
    NovaSingleTask & \textemdash{} & \textemdash{} & \textemdash{} & \textemdash{} & \textemdash{} & \textemdash{} & 0.032 & 79.2\% & \textemdash{} & \textemdash{} & \textemdash{} & \textemdash{} \\
    OpenAIO3 & 0.045 & 30.5\% & 0.035 & 20.2\% & 0.023 & 23.1\% & 0.045 & 20.5\% & 0.037 & 28.2\% & 0.042 & 25.9\% \\
    \bottomrule
    \end{tabular}
    \caption{Inverse Design Results broken down by task type.}
    \label{tab:invdesign}
\end{table}

For the inverse design tasks in \Cref{tab:invdesign}, the 2 or 3 target design appears to be the easiest benchmark, depending on the model; however, there is not a clear trend stating whether more-or-fewer targets is easier. It is not clear why these intermediate task numbers are less difficult than single target design. Our hypothesis is that the individual targets become easier to achieve with increasing target count (either due to profile selection bias or correlation between targets in the real materials we are sampling from), but this is eventually counteracted by having more optimization criteria. More in-depth study is required to deduce why this happens.

\begin{table}[]
    \centering
    \scriptsize
    \begin{tabular}{ccccc}
        \toprule
        Task & \multicolumn{2}{c}{1 View} & \multicolumn{2}{c}{4 View + Code} \\
        Metric & Error & Valid & Error & Valid \\
        Model &  &  &  &  \\
        \midrule
        LLaVAOmniTask & \textbf{0.026} & \textbf{100\%} & 0.023 & \textbf{100\%} \\
        LLaVASingleTask & \textemdash{} & \textemdash{} & \textbf{0.018} & \textbf{100\%} \\
        NovaLite & 0.208 & \textbf{100\%} & 0.192 & \textbf{100\%} \\
        NovaOmniTask & 0.031 & \textbf{100\%} & 0.032 & \textbf{100\%} \\
        NovaSingleTask & \textemdash{} & \textemdash{} & 0.153 & \textbf{100\%} \\
        OpenAIO3 & 0.084 & \textbf{100\%} & 0.071 & \textbf{100\%} \\
        \bottomrule
    \end{tabular}
    \caption{Material Understanding results broken down by task type.}
    \label{tab:understanding}
\end{table}

The expanded material understanding results shown in \Cref{tab:understanding} reveal only that predicting material properties with limited information (a single view), is somewhat more challenging than with an abundance of signal (many views and a MetaDSL representation); this is an unsurprising finding. This discrepancy did lower the overall accuracy of LLaVAOmniTask, but not enough to make a categorical difference.

\subsection{Result Galleries}

We also present randomly\footnote{rejection filtered so that all models had valid outputs for the input, except for inverse design where this was not possible} sampled queries for each task, and visualize their results across models, along with their benchmark metrics. This shows the qualitative differences between the models' performances, while grounding the numeric metrics to make them more understandable.

\begin{figure}
    \centering
    \includegraphics[width=\linewidth]{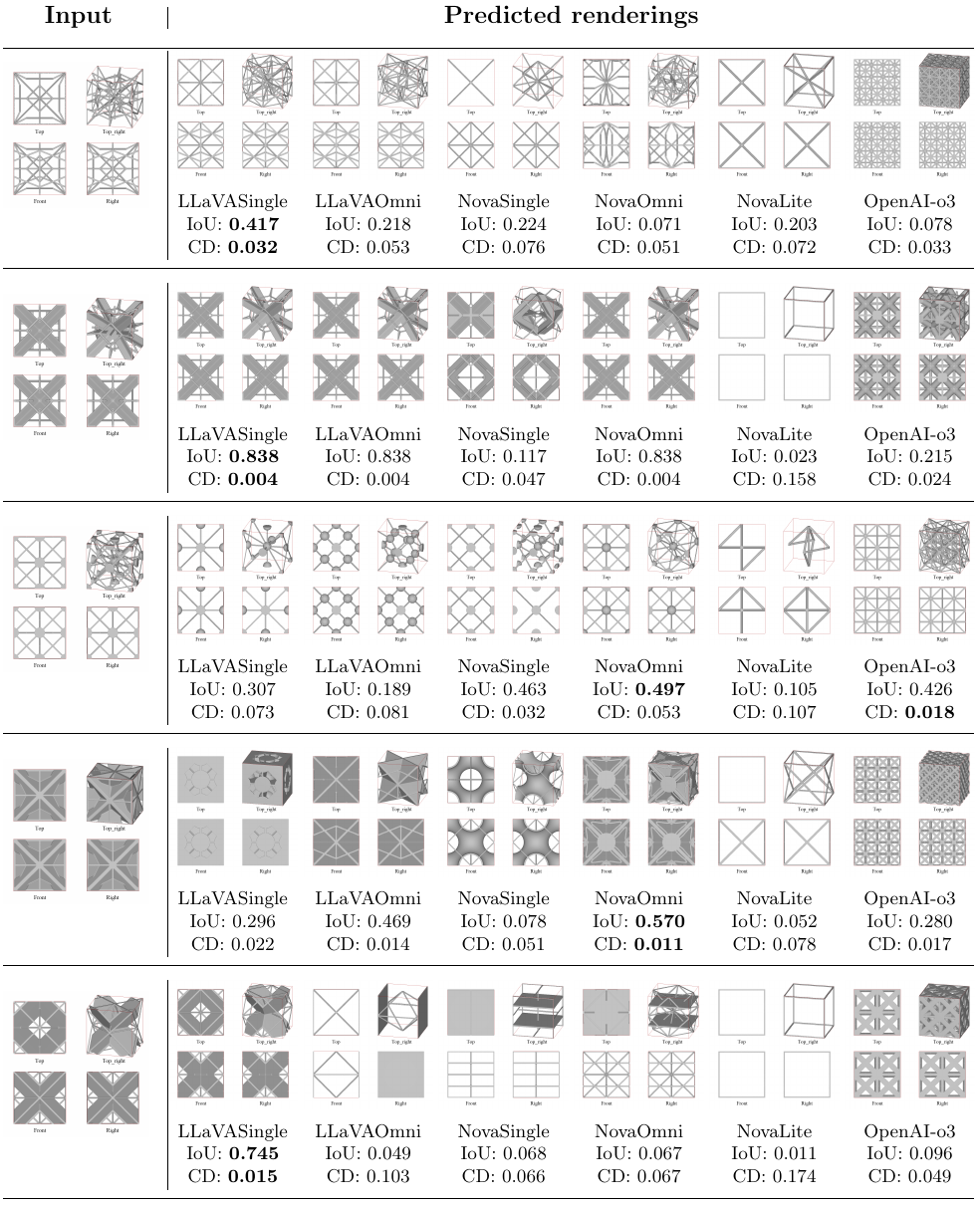}
    \caption{4 View reconstruction results for random test samples by model. Left: the input renders shown to each model. Right: renders of predicted reconstructions.}
    \label{fig:reconstructiongallery}
\end{figure}

\Cref{fig:reconstructiongallery} illustrates reconstruction from 4 viewpoint renders. Of particular interest is the o3 column on the far right. For 4/5 examples, o3 correctly reproduced the basic shape of the side-on views up-to the number of repeats. This suggests that it can correctly build skeletons, but struggles with selecting the correct embedding scale.

\begin{figure}
    \centering
    \includegraphics[width=\linewidth]{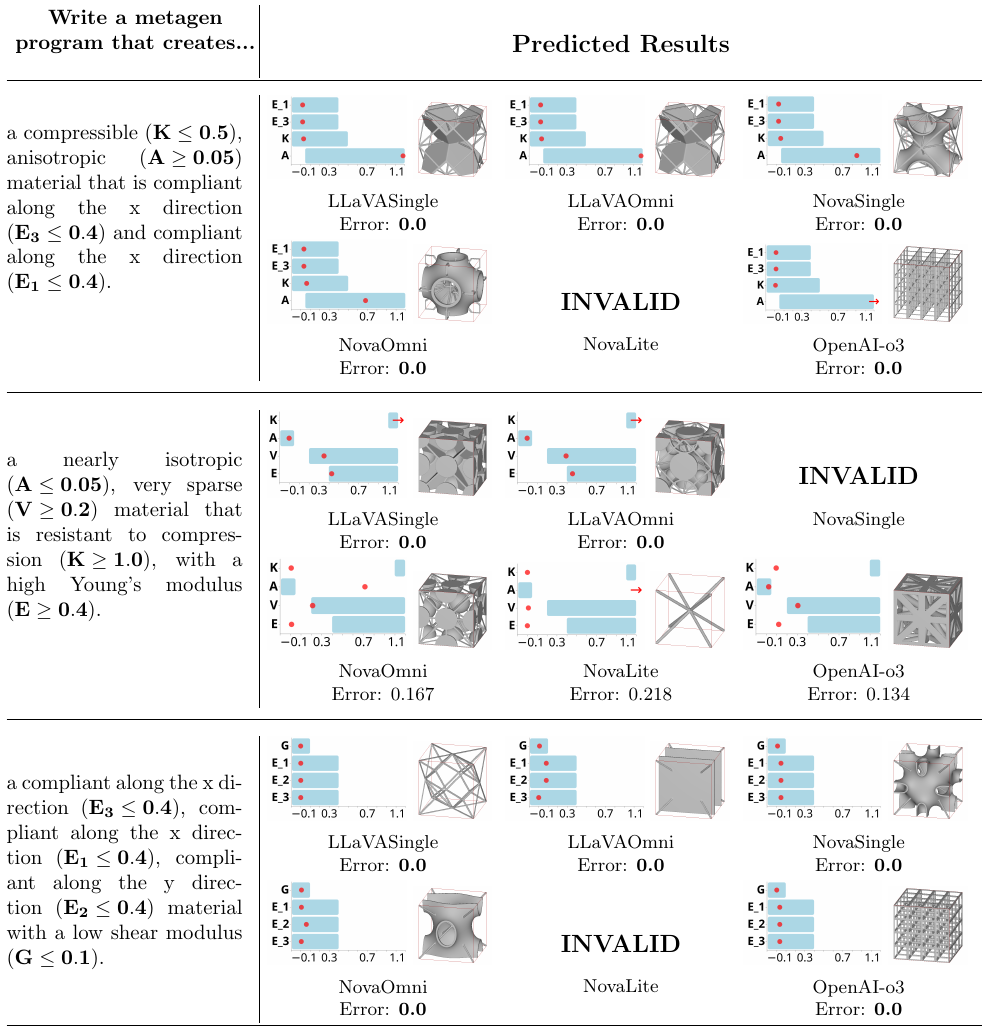}
    \caption{Inverse design results for a random selection of queries. Left: the text query given to each model. Right: paired data showing -- for each model -- an image of the generated structure alongside a property profile comparison. This profile shows the target values/ranges (in blue), versus simulated properties of the predicted materials (in red). Red arrows indicate that the predicted value is beyond the chart boundaries. Some models failed to produce a valid model for certain queries, indicated by the label ``INVALID''.}
    \label{fig:inversedesigngallery}
\end{figure}

\Cref{fig:inversedesigngallery} illustrates material prediction based on specified property requirements. In these examples, the LLaVA models successfully generate materials that meet the given criteria, but other models occasionally generate invalid materials or fail to satisfy the specified requirements.

\begin{figure}
    \centering
    \includegraphics[width=\linewidth]{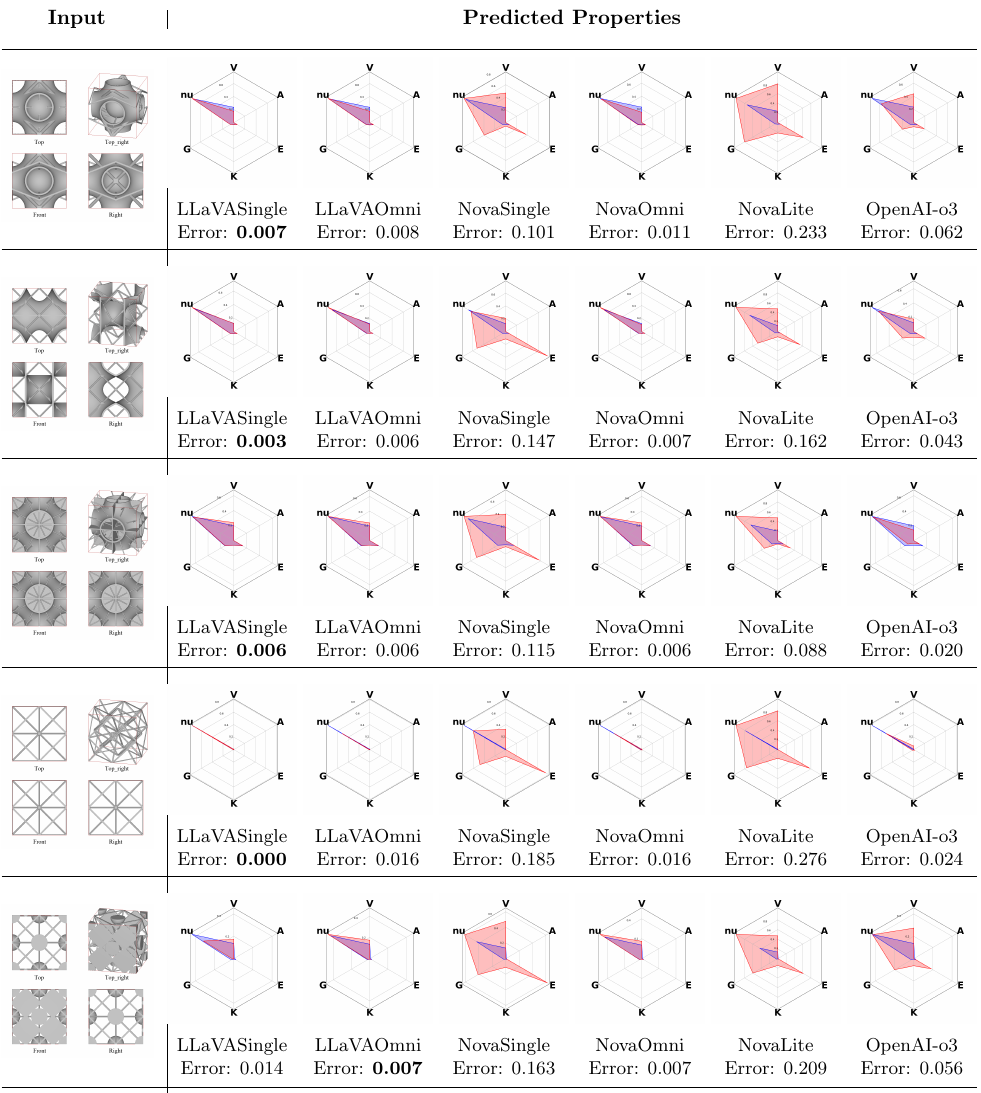}
    \caption{Material property predictions given 4 input views (shown) and the program code (not shown). The radar charts plot the 6 averaged property values (scaled and shifted to always be positive). The blue regions show the ground truth values, while red shows the prediction.}
    \label{fig:understandinggallery}
\end{figure}

\Cref{fig:understandinggallery} illustrates generated materials' predicted versus actual properties. In these examples the LLaVA and OmniTask Nova models do quite well, but single task Nova and untuned models (Novalite and o3) fall behind.

% ==================================
%     Section: Datasets
% ==================================
\section{MetaBench}\label{supp:MetaBench}

\subsection{Intermediate Representation}
\label{sec:app-dataset-intermediate-rep}
Each dataset is given by a set of .jsonl files: one file each for train, validate, and test.
Each line of a .jsonl file describes a single example using a dictionary with the following keys:
\begin{itemize}
    \item \textbf{`task\_type'}: a string identifying the task category; in our case, it is one of \{`reconstruction', `inverse\_design', `material\_understanding'\}.
    \item \textbf{`label'}: unique text label identifying this task entry, using descriptive elements where applicable, such as provided image viewpoints or source files.
    \item \textbf{`source'}: [if applicable] path to the source metamaterial, relative to the database root (and including the leading `/')
    \item \textbf{`data'}: any and all data required to run evaluations, including references for large elements (\eg images, meshes, \etc) and/or directly embedded values. 
    % Although `source` and `data` frequently have overlapping information, we explicitly list the data because it allows us to include test examples that don't reference a particular material (e.g. use an image of a material from literature).
    \item \textbf{`query'}: natural language framing of the question to be provided to an LLM. Any images (or other non-text input) must be specified by reference.
    \item \textbf{`response'}: [optional] an expected response from an LLM that has been asked `query'. This field is permitted to exist for a test example; removal of this information is the responsibility of the LLM-specific formatters, when required.
\end{itemize}

The system prompt has been purposefully excluded, both because it would be very large, and because that is an implementation detail of a predictive model, and not part of the benchmark itself.

\subsection{Task Construction for Inverse Design}
\label{sec:app-dataset-inv-design-prompt-construction}

Inverse design tasks are specified as a collection of target values or bounded-ranges for a subset of material properties, from which we construct a natural-language query that describes that set of targets. Creating these tasks has two stages: selecting a set of targets, and generating an grammatically correct English sentence from those targets.

\paragraph{Property References}
To aid in this process, we generate a reference dictionary with information about each of the $18$ properties, of the following form: 
\begin{lstlisting}[breaklines=True, language=json]
{
'nu': {
    "full_prop_name": "Poisson ratio",
    "alternate_symbols": ["nu_{VRH}"],
    "property_generality": PropertyGenerality.OVERALL,
    "property_type": PropertyType.POISSON_RATIO,
    "dataset_coverage": {
        "min": -0.5,
        "max": 0.5,
        "q1": 0.3,
        "q3": 0.36,
        "densely_populated_ranges": [[0.2, 0.4]]
    },
    "smallest_meaningful_quantization": 0.01,
    "adjective_descriptors":[{"description": f"auxetic", "target_type": TargetType.UPPER_BOUND, "target_value":0}],
    "property_descriptors": [{"description": f"a negative Poisson ratio", "target_type": TargetType.UPPER_BOUND, "target_value":0},
                             {"description": f"a positive Poisson ratio", "target_type": TargetType.LOWER_BOUND, "target_value":0}],
    "verb_descriptors":     [{"description": f"contracts transversely under axial compression", "target_type": TargetType.UPPER_BOUND, "target_value":0},
                             {"description": f"expands transversely under axial compression", "target_type": TargetType.LOWER_BOUND, "target_value":0},
                             {"description": f"contracts in other directions when compressed along one axis", "target_type": TargetType.UPPER_BOUND, "target_value":0},
                             {"description": f"expands in other directions when compressed along one axis", "target_type": TargetType.LOWER_BOUND, "target_value":0},
                             {"description": f"expands transversely under axial elongation", "target_type": TargetType.UPPER_BOUND, "target_value":0},
                             {"description": f"contracts transversely under axial elongation", "target_type": TargetType.LOWER_BOUND, "target_value":0},
                             {"description": f"expands in other directions when stretched along one axis", "target_type": TargetType.UPPER_BOUND, "target_value":0},
                             {"description": f"contracts in other directions when stretched along one axis", "target_type": TargetType.LOWER_BOUND, "target_value":0}]
},
}
\end{lstlisting}
The full listing for all $18$ properties is available in the metagen code provided in the supplement: \verb|metagen/benchmarks_inverse_design.py|.

These entries provide information about the property ranges, dataset coverage, and interesting value breakpoints together with phrases that might be used to request them (\eg, ``auxetic'' implies $\nu < 0$).
All aspects of these reference entries will be used in the following subsections to construct robust, varied and meaningful property queries for different material examples.

\paragraph{Active Property Selection}
For a given structure, we enforce that the ``active'' property subset follows two rules.
First, the active set may only employ the overall values \textit{or} the directional values for any given property -- \eg, if a profile includes measure(s) for Young's modulus, it may either include the overall Young's modulus $E$ \textit{or} one or more of the directional values $\{E_1, E_2, E_3\}$; however, it is not permitted to simultaneously include $E$ and one or more directional variants.
Moreover, a profile is only allowed to use directional variants if it is sufficiently anisiotropic. We chose our anisotropy threshold as $A \ge 0.0025$, based on a manual exploration of the correlation between material spheres and anisotropy values appearing in our dataset.
Subject to these rules, we select the ``active'' subset of properties based on a heuristic that determines the most interesting or salient properties of a given model.

We construct this heuristic score by examining individual properties of a model, and assigning a reward or penalty based on the expected notability of a particular characteristic or combination thereof. 
For example, if a material is near isotropic ($A < 0.0025$), we strongly reward the anisotropy property (so it is likely to end up in the active set) and heavily penalize all directional properties (so they will not be activated, as they are not likely to be notable).
If the material is sufficiently anisotropic, we look at each property with directional variants, then compute pairwise differences between the values (\eg $E_1$ vs. $E_2$). The directional properties are rewarded proportionally to each pairwise difference, so directions with larger discrepancies are more likely to be activated.
Independently, we examine the ratio between the Young's modulus $E$ and the volume fraction $V$ -- if the ratio is high (\ie, the material preserves stiffness with dramatically less material / lighter weight, which is a highly sought after combination), we strongly reward both properties.
Finally, we examine each property in turn, and award additional points if they exhibit values that are extreme and/or underrepresented in our dataset. The reward is proportional to the relative extremity and inversely proportional to representation.

Given these scores, we iteratively select the highest-reward properties that preserve our overall active set rules.
To ensure some variation in our inverse design profiles, we also introduce the opportunity to add randomly chosen properties into our profile: 
after each active set addition from the ranked data, we break the loop with some low probability ($10\%$) and fill the remaining slots with randomly chosen properties that respect the rules relative to our partial active set. 

\paragraph{Active Property Target Selection}
For each active property, we must now select a target value or range. To do this, we evaluate the options present in our reference dictionary, and extract all targets that are satisfied by the material at hand.
We organize these into groups based on value and target type (range, value, lower/upper bound).
Then, we choose the group that offers the tightest bound relative to the current material's property value.
If multiple bound types are associated with the chosen target value, we select a bound type at random.
Finally, we construct a profile with all targets matching the selected value and bound type.
Assuming an example material where the Poisson ratio $\nu = -0.1$, the resulting profile might be as follows:
\begin{lstlisting}[breaklines=True, language=json]
{
    "property": "nu"
    "target_value": 0
    "target_type": "upper_bound"
    "target_descriptions": [
        {
            "description": "auxetic",
            "description_type": "adjective"
        },
        {
            "description": "a negative Poisson ratio",
            "description_type": "noun"
        },
        {
            "description" : "contracts transversely under axial compression",
            "description_type": "verb"
        },
        {
            "description" : "contracts in other directions when compressed along one axis",
            "description_type": "verb"
        },
        {
            "description" : "expands transversely under axial elongation",
            "description_type": "verb"
        },
        {
            "description" : "expands in other directions when stretched along one axis",
            "description_type": "verb"
        }
    ]
}
\end{lstlisting}

\paragraph{Query Construction}

We want to create varied sentence structures to train and test against. To do this, each target type (value, upper bound, or lower bound) and target property has associated with it several descriptive phrases, as shown in the profile above. These phrases are paired with a part of speech (adjective, noun, or verb). As examples ``very dense'' (adjective),  ``contracts in the X direction when the Y direction is stretched'' (verb), or ``a negative Poisson ratio in at least one direction'' (noun). Phrases that do not include numeric targets are accompanied by a parenthetical aside given a target value or range (e.g. ``very dense (V > 0.8).''

We start by randomly selecting one phrase for each target property, binning them by part of speech, then randomizing the order within bins. Adjectives are further randomly split between \emph{front-adjectives} that precede the noun ``material'' (``a very dense material'') and \emph{back-adjectives} that follow it (``a material that is very dense''). We then form a query string by applying the template:
\begin{lstlisting}[breaklines=True]
Write a metagen program that creates [a/an] {front_adjectives} material {back_adjectives} {verbs} {nouns}.    
\end{lstlisting}

The template strings are augmented with part-of-speech appropriate connectors (``that is'', ``with'', ``that'', ``and''), and commas, depending on the parts of number of each part of speech in each position. The pronoun (a/an) as selected based on the first letter of \verb|{front_adjectives}| if there are any, otherwise ``a'' for ``a material''.

\section{Implementation Details}\label{supp:implementation}
LLaVASingleTask and LLaVAOmniTask tune Llama3-LLaVA-Next-8b~\cite{li2024llava, liu2024llavanext} using low-rank adaptation~\cite{hu2022lora}, with with $r=16$ and $\alpha=32$. Models were optimized using AdamW~\cite{loshchilov2017decoupled} with a 1e-5 learning rate and a cosine learning rate scheduler with 0.03 warm-up ratio. SingleTask models were trained on for 7000 iterations on 8 NVIDIA A100 GPUs over approximately 17 hours, while the OmniTask model was trained for just 1 epoch on 8 H200 GPUs over 25 hours due to its significantly larger training set, and for parity with the NovaLiteOmniTask. All LLaVA models were trained with a batch size of 16. During inference, the temperature was set to 0 to ensure deterministic outputs.

For commercial models we primarily used their default settings to avoid excess costs in hyperparameter tuning. NovaSingleTask models were trained on Amazon Bedrock with default settings (2 epochs, learning rate 1e-5, batch size 1, 10 learning rate warmup steps), and NovaOmniTask was trained with the same settings for 1 epoch. NovaSingleTask models trained for 4 hours for reconstruction and material understanding, and 2 hours for inverse design. The NovaOmniTask trained for 24 hours. Default Bedrock parameters were also used at inference time (temperature=0.7, topP=0.9, topK=50). OpenAI's o3 model was queried using the default ``medium'' reasoning level.

\subsection{Training Curves}\label{supp:trainingcurves}
The surprising result that the smaller LLaVA models generally outperformed their much larger Nova counterparts is likely due to the smaller models converging more quickly given the same number of training examples.

\begin{figure}
    \centering
    \includegraphics[width=0.5\linewidth]{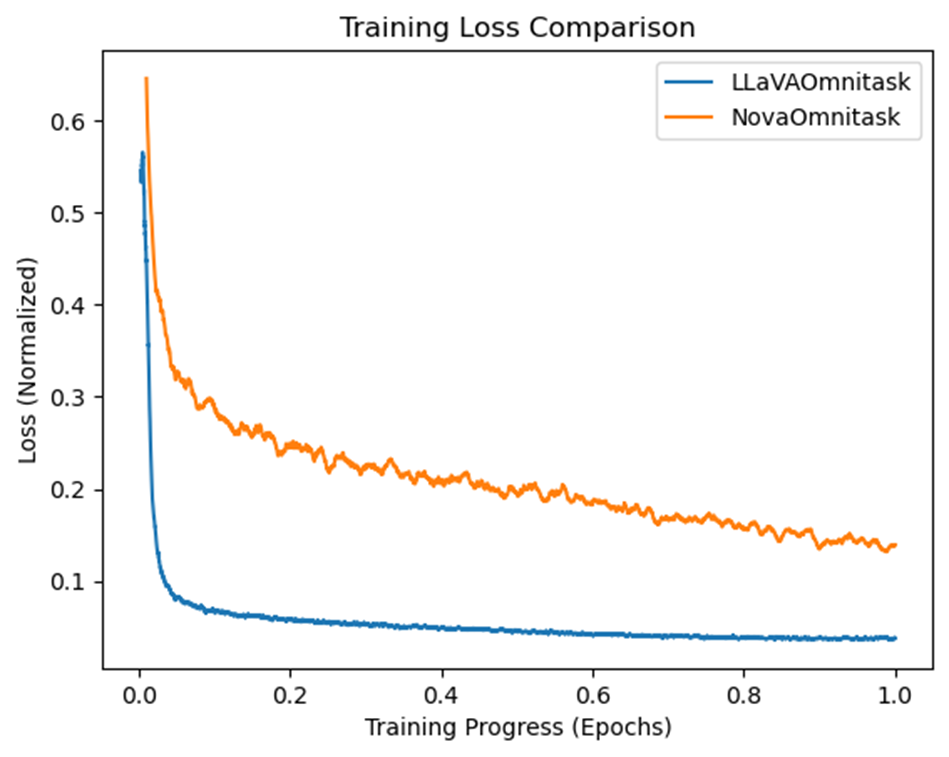}
    \caption{Training loss for LLaVAOmnitask and NovaOmnitask. Losses have been normalized so that starting-loss was 1. The LLaVA model converged very quickly, whereas the Nova loss was still decreasing. Given more training iterations or a steeper learning rate, is is possible that Nova performance would rise to match LLaVA's.}
    \label{fig:trainingloss}
\end{figure}

\subsection{Timing and Costs}
MetaDSL execution and simulation time dominate LLM inference time for material generation. These are highly variable based on the geometric complexity of the generated program, with the majority executing and simulating in 5 minutes or less. MetaAssist generations are on average more time-complex that MetaDB (see \Cref{tab:gensimtimes}. In practice, MetaAssist latencies are much lower because we do not run simulations in the interactive system.

\begin{table}[h!]
    \centering
    \begin{tabular}{c|c|c|c}
         Program Source& Avg. (s) & Median (s) & Std (s)  \\
         \hline
         MetaDB & 181 & 123 & 328 \\
         MetaAssist & 591 & 290 & 746
    \end{tabular}
    \caption{MetaDSL Execution and simulation times for program in MetaDB, and programs generated by MetaAssist using NovaOmni over the MetaBench test set (reconstruction and inverse design).}
    \label{tab:gensimtimes}
\end{table}

Since MetaDSL is quite compact, inference can be performed efficiently with few tokens. The majority of the inference tokens are taken by the common API-description system prompt (\Cref{sec:app-system-prompt}), the cost of which can be amortized by caching. Using NovaOmni (ignoring caching for simplicity), the average MetaBench query used 8730 tokens (8284 input and 446 output). At current API pricing, the average query would cost \$0.0006, and inference for the full test set would cost \$7.11.

% ==================================
%     Section: MetaDSL
% ==================================
\section{Query Templates}
\label{sec:querytemplates}

For training models and running inference, we used prompt templates and inserted details for each specific query. In the following templates, \verb|<[ ... ]>| is used as a delimiter to denote the inclusion of an image.

\subsection{Universal System Prompt}
\label{sec:app-system-prompt}
For consistency, every example was provided with a common system prompt that describes the Metagen DSL, explains the material properties and rendered views we have in our dataset, and describes the basic task categories.

\begin{lstlisting}[breaklines=True]
You are an expert metamaterials assistant that generates and analyzes cellular metamaterial designs based on material properties, images, and programatic definitions in the Metagen metamaterial DSL.


# Procedural Description in a Metamaterial DSL:

{api_description}

# Material Analysis:
You can analyze the density, anisotropy, and elasticity properties of metamaterials. All metamaterials are assumed to be constucted from an isotropic base material with Poisson's ratio nu = 0.45.
The Young's Modulus of this base material is not specified, instead, the elastic moduli of the metamaterials -- Young's Modulus (E), Bulk Modulus (K), and Shear Modulus (G), are expressed relative to the base material Young's modulus (E_base). This means, for example, that relative Young's Moduli can range from 0 to 1. The material properties you can analyze are:

- E: Young's Modulus, Voigt-Reuss-Hill (VRH) average, relative to E_base
- E_1,E_2,E_3: Directional Young's Moduli, relative to E_base
- G: Shear Modulus (VRH average), relative to E_base
- G_23,G_13,G_12: Directional Shear Moduli, relative to E_base
- nu: Poisson ratio (VRH average)
- nu_12, nu_13, nu_23, nu_21, nu_31, nu_32: Directional Poisson ratios
- K: Bulk modulus (VRH average), relative to E_base
- A: Anisotropy (universal anisotropy index)
- V: Volume Fraction

# Material Images:

Images of metamaterials depict a base cell of the material rendered from four viewpoints:

- from the top
- from the front side
- from the right side
- from an angle at the upper-front-right

# Tasks:

You will be asked to perform several kinds of tasks:

- Reconstruction: from one or more images of a target material, reconstruct a Metagen program that generates the metamaterial in the images.
- Inverse Design: from a description of the properties of a desired materials, write a Metagen program that creates a metamaterial with those properties.
- Material Understanding: from images of a metamaterial and/or a Metagen program, analyze a material and predict its properties.
\end{lstlisting}

\subsection{MetaDSL API}
\label{sec:app-metadsl-api}
The Metagen language description (inserted as the \verb|api_description| in the system prompt above) is as follows:

\begin{lstlisting}[breaklines=True]

Programs in Metagen are built in two stages: one that creates local geometric structure, and a second that patterns this structure throughout space. Each of these is further broken down into subparts.


==================================
    API description (Boilerplate)
==================================
Each program is given as a python file (.py).
This program must import the metagen package and define a function called "make_structure()", which returns the final Structure object defined by the program. 
If parameters are present in make_structure(), they MUST have a default value.
Specifically, the file structure is as follows: 


from metagen import *

def make_structure(...) -> Structure:
    <content>



==================================
    DSL description
==================================

======= Skeleton Creation ========
vertex(cpEntity, t)
    @description:
        Create a new vertex. This vertex is defined relative to its containing convex polytope (CP). It will only have an embedding in R3 once the CP has been embedded.
    @params:
        cpEntity    - an entity of a convex polytope (CP), referenced by the entity names.
        t           - [OPTIONAL] list of floats in range [0,1], used to interpolate to a specific position on the cpEntity.
                        If cpEntity is a corner, t is ignored.
                        If cpEntity is an edge, t must contain exactly 1 value. t is used for linear interpolation between the endpoints of cpEntity.
                        If cpEntity is a face, t must contain exactly 2 values. If cpEntity is a triangular face, t is used to interpolate via barycentric coordinates. If cpEntity is a quad face, bilinear interpolation is used.
                        
                        If the optional interpolant t is omitted for a non-corner entity, the returned point will be at the midpoint (for edge) or the centroid (for face) of the entity. Semantically, we encourage that t be excluded (1) if the structure would be invalid given a different non-midpoint t, or (2) if the structure would remain unchanged in the presence a different t (e.g., in the case of a conjugate TPMS, where only the entity selection matters).
    @returns:
        vertex      - the new vertex object 
    @example_usage:
        v0 = vertex(cuboid.edges.BACK_RIGHT, [0.5])
        v1 = vertex(cuboid.edges.TOP_LEFT)


Polyline(ordered_verts)
    @description:
        Creates a piecewise-linear path along the ordered input vertices. All vertices must be referenced to the same CP (e.g., all relative to cuboid entities). The resulting path will remain a polyline in any structures that include it.
    @params:
        ordered_verts   - a list of vertices, in the order you'd like them to be traversed. A closed loop may be created by repeating the zeroth element at the end of the list. No other vertex may be repeated. Only simple paths are permitted.
    @returns:
        polyline        - the new polyline object
    @example_usage:
        p0 = Polyline([v2, v3])
        p0 = Polyline([v0, v1, v2, v3, v4, v5, v0])


Curve(ordered_verts)
    @description:
        Creates a path along the ordered input vertices. This path will be smoothed at a later stage (e.g., to a Bezier curve), depending on the lifting procedures that are chosen. All input vertices must be referenced to the same CP (e.g., all relative to cuboid entities). 
    @params:
        ordered_verts   - a list of vertices, in the order you'd like them to be traversed. A closed loop may be created by repeating the zeroth element at the end of the list. No other vertex may be repeated. Only simple paths are permitted.
    @returns:
        curve           - the new curve object
    @example_usage:
        c0 = Curve([v2, v3])
        c0 = Curve([v0, v1, v2, v3, v4, v5, v0])

skeleton(entities)
    @description:
        Combines a set of vertices OR polylines/curves into a larger structure, over which additional information can be inferred. For example, within a skeleton, multiple open polylines/curves may string together to create a closed loop, a branched path, or a set of disconnected components.
    @params:
        entities        - a list of entities (vertices or polylines/curves) to be combined. A given skeleton must only have entities with the same dimension -- that is, it must consist of all points or all polylines/curves.
    @returns:
        skeleton        - the new skeleton object
    @example_usage:
        skel = skeleton([curve0, polyline1, curve2, polyline3])
        skel = skeleton([v0])


======= Lifting Procedures ========
UniformBeams(skel, thickness)
    @description:
        Procedure to lift the input skeleton to a 3D volumetric structure by instantiating a beam of the given thickness centered along each polyline/curve of the input skeleton.
    @requirements:
        The skeleton must contain only polylines and/or curves. The skeleton must not contain any standalone vertices.
    @params:
        skel            - the skeleton to lift
        thickness       - the diameter of the beams
    @returns:
        liftProc        - the lifted skeleton
    @example_usage:
        liftProcedure = UniformBeams(skel, 0.03)

SpatiallyVaryingBeams(skel, thicknessProfile)
    @description:
        Procedure to lift the input skeleton to a 3D volumetric structure by instantiating a beam of the given spatially-varying thickness profile centered along each polyline/curve of the input skeleton.
    @requirements:
        The skeleton must contain only polylines and/or curves. The skeleton must not contain any standalone vertices.
    @params:
        skel            - the skeleton to lift
        thicknessProfile- specifications for the diameter of the beams along each polyline/curve. Given as a list[list[floats]], where the each of the n inner lists gives the information for a single sample point along the polyline/curve. The first element in each inner list provides a position parameter t\\in[0,1] along the polyline/curve, and the second element specifies the thickness of the beam at position t
    @returns:
        liftProc        - the lifted skeleton
    @example_usage:
        liftProcedure = SpatiallyVaryingBeams(skel, 0.03)

UniformDirectShell(skel, thickness)
    @description:
        Procedure to lift the input skeleton to a 3D volumetric structure by inferring a surface that conforms to the boundary provided by the input skeleton. The surface is given by a simple thin shell model: the resulting surface is incident on the provided boundary while minimizing a weighted sum of bending and stretching energies. The boundary is fixed, though it may be constructed with a mix of polylines and curves (which are first interpolated into a spline, then fixed as part of the boundary). The skeleton must contain a single closed loop composed of one or more polylines and/or curves. The skeleton must not contain any standalone vertices.
    @requirements:

    @params:
        skel            - the skeleton to lift
        thickness       - the thickness of the shell. The final offset is thickness/2 to each side of the inferred surface.
    @returns:
        liftProc        - the lifted skeleton
    @example_usage:
        liftProcedure = UniformDirectShell(skel, 0.1)

UniformTPMSShellViaConjugation(skel, thickness)
    @description:
        Procedure to lift the input skeleton to a 3D volumetric structure by inferring a triply periodic minimal surface (TPMS) that conforms to the boundary constraints provided by the input skeleton. The surface is computed via the conjugate surface construction method. 
    @requirements: 
        The skeleton must contain a single closed loop composed of one or more polylines and/or curves. The skeleton must not contain any standalone vertices.
        Each vertex in the polylines/curves must live on a CP edge.
        Adjacent vertices must have a shared face. 
        The loop must touch every face of the CP at least once.
        If the CP has N faces, the loop must contain at least N vertices.
    @params:
        skel            - the skeleton to lift
        thickness       - the thickness of the shell. The final offset is thickness/2 to each side of the inferred surface.
    @returns:
        liftProc        - the lifted skeleton
    @example_usage:
        liftProcedure = UniformTPMSShellViaConjugation(skel, 0.03)

UniformTPMSShellViaMixedMinimal(skel, thickness)
    @description:
        Procedure to lift the input skeleton to a 3D volumetric structure by inferring a triply periodic minimal surface (TPMS) that conforms to the boundary constraints provided by the input skeleton. The surface is computed via mean curvature flow. All polyline boundary regions are considered fixed, but any curved regions may slide within their respective planes in order to reduce surface curvature during the solve.
    @requirements: 
        The skeleton must contain a single closed loop composed of one or more polylines and/or curves. The skeleton must not contain any standalone vertices.
        Each vertex in the polylines/curves must live on a CP edge.
        Adjacent vertices must have a shared face. 
    @params:
        skel            - the skeleton to lift
        thickness       - the thickness of the shell. The final offset is thickness/2 to each side of the inferred surface.
    @returns:
        liftProc        - the lifted skeleton
    @example_usage:
        liftProcedure = UniformTPMSShellViaMixedMinimal(skel, 0.03)

Spheres(skel, thickness)
    @description:
        Procedure to lift the input skeleton to a 3D volumetric structure by instantiating a sphere of the given radius centered at vertex p, for each vertex in the skeleton.
    @requirements:
        The skeleton must only contain standalone vertices; no polylines or curves can be used.
    @params:
        skel            - the skeleton to lift
        thickness       - the sphere radius 
    @returns:
        liftProc        - the lifted skeleton
    @example_usage:
        s_lift = Spheres(skel, 0.25)


======= Tile Creation ========
Tile(lifted_skeletons, embedding)
    @description:
        Procedure to embed a copy of the skeleton in R^3 using the provided embedding information. The embedding information can be computed by calling the "embed" method of the relevant CP. 
    @requirements:
        The embedding information must correspond to the same CP against which the vertices were defined. For example, if the vertices are defined relative to the cuboid, you must use the cuboid.embed() method.
    @params:
        lifted_skeletons- a list of lifted skeleton entities to embed in R^3. All entities must reside in the same CP type, and this type must have N corners.
        embedding       - information about how to embed the CP and its relative skeletons within R^3. Obtained using the CP's embed() method
    @returns:
        tile            - the new tile object
    @example_usage:
        embedding = cuboid.embed(side_len, side_len, side_len, cornerAtAABBMin=cuboid.corners.FRONT_BOTTOM_LEFT)
        s_tile = Tile([beams, shell], embedding)


======= Patterning Procedures ========
TetFullMirror()
    @description:
        Procedure which uses only mirrors to duplicate a tet-based tile such that it partitions R^3
    @params:
        N/A
    @returns:
        pat     - the patterning procedure
    @example_usage:
        pat = TetFullMirror()

TriPrismFullMirror()
    @description:
        Procedure which uses only mirrors to duplicate a triangular prism-based tile such that it partitions R^3
    @params:
        N/A
    @returns:
        pat     - the patterning procedure
    @example_usage:
        pat = TriPrismFullMirror()

CuboidFullMirror()
    @description:
        Procedure which uses only mirrors to duplicate an axis-aligned cuboid tile such that it fills a unit cube,  such that it partitions R^3. Eligible cuboid CPs must be such that all dimensions are 1/(2^k) for some positive integer k.
    @params:
        N/A
    @returns:
        pat     - the patterning procedure
    @example_usage:
        pat = CuboidFullMirror()

Identity()
    @description:
        No-op patterning procedure.
    @params:
        N/A
    @returns:
        pat     - the patterning procedure
    @example_usage:
        pat = Identity()

Custom(patternOp)
    @description:
        Environment used to compose a custom patterning procedure. Currently only implemented for the Cuboid CP.
    @params:
        patternOp- outermost pattern operation in the composition
    @returns:
        pat     - the complete patterning procedure
    @example_usage:
        pat = Custom(Rotate180([cuboid.edges.BACK_RIGHT, cuboid.edges.BACK_LEFT], True,
                        Rotate180([cuboid.edges.TOP_RIGHT], True)))

Mirror(entity, doCopy, patternOp)
    @description:
        Pattern operation specifying a mirror over the provided CP entity, which must be a CP Face. Can only be used inside of a Custom patterning environment.
    @params:
        entity   - CP Face that serves as the mirror plane. 
        doCopy   - boolean. When True, applies the operation to a copy of the input, such that the original and the transformed copy persist. When False, directly transforms the input.
        patternOp- [OPTIONAL] outermost pattern operation in the sub-composition, if any
    @returns:
        pat      - the composed patterning procedure, which may be used as is (within the Custom environment), or as the input for further composition
    @example_usage:
        pat = Custom(Mirror(cuboid.faces.TOP, True, 
                        Mirror(cuboid.faces.LEFT, True)))

Rotate180(entities, doCopy, patternOp)
    @description:
        Pattern operation specifying a 180 degree rotation about the provided CP entity. Can only be used inside of a Custom patterning environment.
    @params:
        entities - List of CP entities, which define the axis about which to rotate. If a single entity is provided, it must be a CP Edge. If multiple entities, they will be used to define a new entity that spans them. For example, if you provide two corners, the axis will go from one to the other. If you provide two CP Edges, the axis will reach from the midpoint of one to the midpoint of the other.
        doCopy   - boolean. When True, applies the operation to a copy of the input, such that the original and the transformed copy persist. When False, directly transforms the input.
        patternOp- [OPTIONAL] outermost pattern operation in the sub-composition, if any
    @returns:
        pat      - the composed patterning procedure, which may be used as is (within the Custom environment), or as the input for further composition
    @example_usage:
        pat = Custom(Rotate180([cuboid.edges.FRONT_LEFT, cuboid.edges.FRONT_RIGHT], True))

Translate(fromEntity, toEntity, doCopy, patternOp)
    @description:
        Pattern operation specifying a translation that effectively moves the fromEntity to the targetEntity. Can only be used inside of a Custom patterning environment.
    @params:
        fromEntity- CP Entity that serves as the origin of the translation vector. Currently only implemented for a CP Face.
        toEntity  - CP Entity that serves as the target of the translation vector. Currently only implemented for a CP Face.
        doCopy   - boolean. When True, applies the operation to a copy of the input, such that the original and the transformed copy persist. When False, directly transforms the input.
        patternOp- [OPTIONAL] outermost pattern operation in the sub-composition, if any
    @returns:
        pat      - the composed patterning procedure, which may be used as is (within the Custom environment), or as the input for further composition
    @example_usage:
        gridPat = Custom(Translate(cuboid.faces.LEFT, cuboid.faces.RIGHT, True,
                                Translate(cuboid.faces.FRONT, cuboid.faces.BACK, True)))


======= Structure Procedures ========
Structure(tile, pattern)
    @description:
        Combines local tile information (containing lifted skeletons) with the global patterning procedure to generate a complete metamaterial.
    @params:
        tile            - the tile object, which has (by construction) already been embedded in 3D space, along with all lifted skeletons it contains.
        pattern         - the patterning sequence to apply to extend this tile throughout space
    @returns:
        structure       - the new structure object
    @example_usage:
        obj = Structure(tile, pat)

Union(A, B)
    @description:
        Constructive solid geometry Boolean operation that computes the union of two input structures. The output of Union(A,B) is identical to Union(B,A)
    @params:
        A               - the first Structure to be unioned. This may be the output of Structure, Union, Subtract, or Intersect
        B               - the second Structure to be unioned. This may be the output of Structure, Union, Subtract, or Intersect
    @returns:
        structure       - the new structure object containing union(A,B)
    @example_usage:
        final_obj = Union(schwarzP_obj, Union(sphere_obj, beam_obj))

Subtract(A, B)
    @description:
        Constructive solid geometry Boolean operation that computes the difference (A - B) of two input structures. The relative input order is critical.
    @params:
        A               - the first Structure, from which B will be subtracted. This may be the output of Structure, Union, Subtract, or Intersect
        B               - the second Structure, to be subtracted from A. This may be the output of Structure, Union, Subtract, or Intersect
    @returns:
        structure       - the new structure object containing (A - B)
    @example_usage:
        final_obj = Subtract(c_obj, s_obj)

Intersect(A, B)
    @description:
        Constructive solid geometry Boolean operation that computes the intersection of two input structures, A and B. 
    @params:
        A               - the first Structure, which may be the output of Structure, Union, Subtract, or Intersect
        B               - the second Structure, which may be the output of Structure, Union, Subtract, or Intersect
    @returns:
        structure       - the new structure object containing the intersection of A and B
    @example_usage:
        final_obj = Intersect(c_obj, s_obj)




==================================
    Prebuilt Convex Polytopes
==================================
There are 3 prebuilt convex polytopes (CP) available for use: cuboid, triPrism, and tet. Each CP comprises a set of Entities, namely faces, edges and corners. 
For convenience, each individual entity can be referenced using the pattern <CP>.<entity_type>.<ENTITY_NAME>. 
For example, you can select a particular edge of the cuboid with the notation cuboid.edges.BOTTOM_RIGHT.
Each CP also has an embed() method which returns all necessary information to embed the CP within R^3.

The full list of entities and embed() method signatures for our predefined CPs are as follows:

tet.corners.{   BOTTOM_RIGHT,
                BOTTOM_LEFT,
                TOP_BACK,
                BOTTOM_BACK
            }
tet.edges.  {   BOTTOM_FRONT,
                TOP_LEFT,
                BACK,
                BOTTOM_RIGHT,
                TOP_RIGHT,
                BOTTOM_LEFT
            }
tet.faces.  {   BOTTOM,
                TOP,
                RIGHT,
                LEFT
            }
tet.embed(bounding_box_side_length)
    @description:
        Constructs the information required to embed the tet CP in R^3
    @params:
        bounding_box_side_length- length of axis-aligned bounding box containing the tet. Float in range [0,1]. Must be 1/2^k for some integer k
    @returns:
        embedding      - the embedding information. Specifically, the position in R^3 of all the CP corners. 
    @example_usage:
        side_len = 0.5 / num_tiling_unit_repeats_per_dim
        embedding = tet.embed(side_len)


triPrism.corners.{FRONT_BOTTOM_LEFT,
                FRONT_TOP,
                FRONT_BOTTOM_RIGHT,
                BACK_BOTTOM_LEFT,
                BACK_TOP,
                BACK_BOTTOM_RIGHT
            }
triPrism.edges.{FRONT_LEFT,
                FRONT_RIGHT,
                FRONT_BOTTOM,
                BACK_LEFT,
                BACK_RIGHT,
                BACK_BOTTOM,
                BOTTOM_LEFT,
                TOP,
                BOTTOM_RIGHT
            }
triPrism.faces.{FRONT_TRI,
                BACK_TRI,
                LEFT_QUAD,
                RIGHT_QUAD,
                BOTTOM_QUAD
            }
triPrism.embed(bounding_box_side_length)
    @description:
        Constructs the information required to embed the triangular prism CP in R^3
    @params:
        bounding_box_side_length - length of axis-aligned bounding box containing the triangular prism. Float in range [0,1]. Must be 1/2^k for some integer k
    @returns:
        embedding      - the embedding information. Specifically, the position in R^3 of all the CP corners. 
    @example_usage:
        side_len = 0.5 / num_tiling_unit_repeats_per_dim
        embedding = triPrism.embed(side_len)


cuboid.corners.{FRONT_BOTTOM_LEFT,
                FRONT_BOTTOM_RIGHT,
                FRONT_TOP_LEFT,
                FRONT_TOP_RIGHT,
                BACK_BOTTOM_LEFT,
                BACK_BOTTOM_RIGHT,
                BACK_TOP_LEFT,
                BACK_TOP_RIGHT
            }
cuboid.edges.{  FRONT_BOTTOM,
                FRONT_LEFT,
                FRONT_TOP,
                FRONT_RIGHT,
                BACK_BOTTOM,
                BACK_LEFT,
                BACK_TOP,
                BACK_RIGHT,
                BOTTOM_LEFT,
                TOP_LEFT,
                TOP_RIGHT,
                BOTTOM_RIGHT
            }
cuboid.faces.{  FRONT,
                BACK,
                TOP,
                BOTTOM,
                LEFT,
                RIGHT
            }
            
cuboid.embed(width, height, depth, cornerAtMinPt)
    @description:
        Constructs the information required to embed the cuboid CP in R^3
    @params:
        width          - length of cuboid side from left to right. float in range [0,1]. Must be 1/2^k for some integer k
        height         - length of cuboid side from top to bottom. float in range [0,1]. Must be 1/2^k for some integer k
        depth          - length of cuboid side from front to back. float in range [0,1]. Must be 1/2^k for some integer k
        cornerAtMinPt  - CP corner entity (e.g., cuboid.corners.FRONT_BOTTOM_LEFT) that should be collocated with the cuboid's minimum position in R^3
    @returns:
        embedding      - the embedding information. Specifically, the position in R^3 of all the CP corners. 
    @example_usage:
        side_len = 0.5 / num_tiling_unit_repeats_per_dim
        embedding = cuboid.embed(side_len, side_len, side_len, cornerAtAABBMin=cuboid.corners.FRONT_BOTTOM_LEFT)

cuboid.embed_via_minmax(aabb_min_pt, aabb_max_pt, cornerAtMinPt)
    @description:
        Constructs the information required to embed the cuboid CP in R^3
    @params:
        aabb_min_pt    - Minimum point of the cuboid, in R^3. Given as a list of length 3, where each component must be a float in range [0,1], with 1/2^k for some integer k
        aabb_max_pt    - Maximum point of the cuboid, in R^3. Given as a list of length 3, where each component must be a float in range [0,1], with 1/2^k for some integer k
        cornerAtMinPt  - CP corner entity (e.g., cuboid.corners.FRONT_BOTTOM_LEFT) that should be collocated with the cuboid's minimum position in R^3
    @returns:
        embedding      - the embedding information. Specifically, the position in R^3 of all the CP corners. 
    @example_usage:
        side_len = 0.5 / num_tiling_unit_repeats_per_dim
        embedding = cuboid.embed([0,0,0], [side_len, side_len, side_len], cuboid.corners.BACK_BOTTOM_RIGHT)

\end{lstlisting}

\paragraph{API Errata}
The API description listed in this section is the exact version we used to train all models in MetaBench. This differs slightly from the released version, which corrects two mistakes that were identified at a later stage:
\begin{itemize}
    \item \verb|cuboid.embed()|: the original description (above) listed a parameter \verb|cornerAtMinPt| in both the signature line and the \verb|@params| listing. However, the \verb|@example_usage| showed the parameter as \verb|cornerAtAABBMin|. The latter is correct, and reflects an update made in the code independently of the documentation. The released API description consistently shows the correct parameter name,  \verb|cornerAtAABBMin|.
    \item \verb|cuboid.embed_via_minmax()|: the \verb|@example_usage| field of the original description (above) erroneously lists the \verb|cuboid.embed()| function with the inputs of the intended function, \verb|cuboid.embed_via_minmax()|. None of the parameters were updated, as they are all correct in the original description above. Only the erroneous function call was corrected in the released version (\verb|cuboid.embed()| $\rightarrow$ \verb|cuboid.embed_via_minmax()|).
\end{itemize}

These mistakes did not cause any observable issue in the trained model output, as the (correctly expressed) training data overrode the error in our API description. 
However, this did cause an issue for zero shot experiments (which ultimately revealed the bug). 
All zero shot results reported in the paper reflect the results using the \textit{updated} version of our API, where the difference relative to the listing above constitutes exactly the two changes discussed here.

To ensure that this API description would not derail otherwise successful program outputs (and to mitigate confusion between the two very similar keywords across functions), we added an optional keyword argument to the signature of both affected functions, such that either keyword (or no keyword, as in a positional argument) is permissible.
Thus, either API description is suitable; however, we release the corrected version to prevent issues and reduce confusion moving forward.

\subsection{Reconstruction}
\label{sec:querytemplates_reconstruction}

Reconstruction tasks can have any combination of one to four views. Here we only reproduced the 4 view template; the others have the irrelevant lines removed.
\begin{lstlisting}[breaklines=True]
# Task:
Analyze these views of a metamaterial, then generate a metamaterial DSL procedure to reproduce it.

# Inputs:
**Rendered Views:**
Top: <[{top}]>
Front: <[{front}]>
Right: <[{right}]>
Angled (Front-Top-Right): <[{top_right}]>

# Output Format:
Generate a Metagen program within a python code block:

```python
from metagen import *

def make_structure(...) -> Structure:
    ...
```

\end{lstlisting}

\subsection{Inverse Design}
\label{sec:querytemplates_inverse}

\begin{lstlisting}[breaklines=True]
# Task:
Write a metagen program that creates {query_target}.

# Output Format:
Generate a Metagen program within a python code block:

```python
from metagen import *

def make_structure(...) -> Structure:
    ...
```
\end{lstlisting}

\subsection{Material Understanding}
\label{sec:querytemplates_understanding}

Single View:
\begin{lstlisting}[breaklines=True]
# Task:
Analyze these views of a metamaterial, and predict its material properties.

# Inputs:

**Rendered View:**

- Angled (Front-Top-Right): <[{top_right}]>

# Output Format:

Output a json object, delimited by ```json ```, where the keys are material property names, and the values are the predicted material properties. Predict these properties (keys):
- "A" : Anisotropy (universal anisotropy index)
- "E" : Young's Modulus relative to E_base
- "K" : Bulk modulus relative to E_base
- "G": Shear modulus relative to E_base
- "nu": Isotropic Poisson ratio
- "V" : Relative Density (Volume Fraction)
\end{lstlisting}

Multiview + Code:
\begin{lstlisting}[breaklines=True]
# Task:
Analyze these views of a metamaterial, and the Metagen program, and predict its material properties.

# Inputs:

**Metagen Program:**

{code}

**Rendered Views:**
- Top: <[{top}]>
- Front: <[{front}]>
- Right: <[{right}]>
- Angled (Front-Top-Right): <[{top_right}]>

# Output Format:

Output a json object, delimited by ```json ```, where the keys are material property names, and the values are the predicted material properties. Predict these properties (keys):
- "A" : Anisotropy (universal anisotropy index)
- "E" : Young's Modulus relative to E_base
- "K" : Bulk modulus relative to E_base
- "G": Shear modulus relative to E_base
- "nu": Isotropic Poisson ratio
- "V" : Relative Density (Volume Fraction)
\end{lstlisting}

\end{document}